\pdfoutput=1

\documentclass[11pt]{article}

\usepackage[]{acl}

\usepackage{times}
\usepackage{latexsym}

\usepackage[T1]{fontenc}

\usepackage[utf8]{inputenc}

\usepackage{color}
\usepackage{tabularx}
\usepackage{tabu}
\usepackage{booktabs}
\usepackage{multirow}

\usepackage{graphicx} 
\usepackage{float}
\usepackage{subfigure} 
\usepackage{amssymb}

\usepackage{hyperref}
\usepackage{tablefootnote}
\usepackage{xspace}
\usepackage{tcolorbox}

\usepackage{float}
\usepackage{amsmath}
\usepackage{bm}
\usepackage{algorithmicx}
\usepackage{algorithm}
\usepackage{algpseudocode}

\usepackage{arydshln}

\usepackage{amssymb}
\usepackage{pifont}
\newcommand{\cmark}{\ding{51}}%
\usepackage{enumitem}
\usepackage{fontawesome5}
\usepackage{microtype}

\newcommand{\ours}{\textsc{LimitGen}\xspace}
\newcommand{\oursone}{\textsc{LimitGen}-Syn\xspace}
\newcommand{\ourstwo}{\textsc{LimitGen}-Human\xspace}
\newcommand{\eg}{\hbox{\emph{e.g.,}}\xspace}
\newcommand{\ie}{\hbox{\emph{i.e.,}}\xspace}

\newcommand{\huggingface}{\raisebox{-1.5pt}{\includegraphics[height=1.05em]{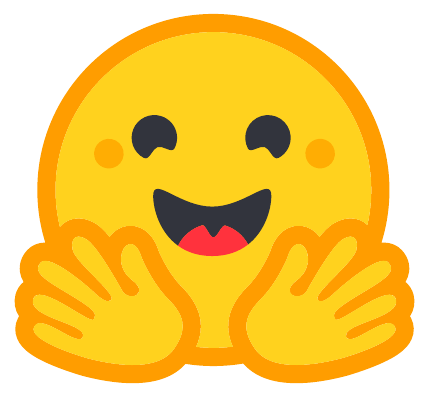}}\xspace}
\newcommand{\github}{\raisebox{-1.5pt}{\includegraphics[height=1.05em]{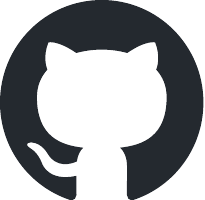}}\xspace}

\definecolor{brightmaroon}{rgb}{0.76, 0.13, 0.28}

\definecolor{YaleBlue}{RGB}{16, 42, 86}
\definecolor{TATABlue}{RGB}{1, 126, 199}

\newcommand{\Yale}{\hspace{.1em}^{\textcolor{YaleBlue}{\boldsymbol{Y}}}}
\newcommand{\Tata}{\hspace{.1em}^{\textcolor{TATABlue}{\boldsymbol{T}}}}

\newcommand{\rqone}{How well do LLM-based systems perform in identifying limitations within scientific research?}
\newcommand{\rqtwo}{Can RAG enhance LLMs' ability to identify limitations and provide constructive suggestions?}

\newcommand{\rqthree}{
How can this research be applied in real-world scenarios to assist human researchers in improving their work?}

\title{Can LLMs Identify Critical Limitations within Scientific Research?\\ 
A Systematic Evaluation on AI Research Papers}

\author{
Zhijian Xu\thanks{~~Equal Contributions.}~$\Yale$ \quad Yilun Zhao$^{*}$$\Yale$ \quad Manasi Patwardhan$\Tata$ \\\bf{Lovekesh Vig$\Tata$ \quad Arman Cohan$\Yale$} \vspace{4pt}\\
$\Yale$~Yale University \quad $\Tata$~TCS Research
\vspace{10pt}
}

\begin{document}
\maketitle


\begin{abstract}
Peer review is fundamental to scientific research, but the growing volume of publications has intensified the challenges of this expertise-intensive process. While LLMs show promise in various scientific tasks, their potential to assist with peer review, particularly in identifying paper limitations, remains understudied. We first present a comprehensive taxonomy of limitation types in scientific research, with a focus on AI. Guided by this taxonomy,  for studying limitations, we present \ours, the first comprehensive benchmark for evaluating LLMs' capability to support early-stage feedback and complement human peer review. Our benchmark consists of two subsets: \oursone, a synthetic dataset carefully created through controlled perturbations of high-quality papers, and \ourstwo, a collection of real human-written limitations. To improve the ability of LLM systems to identify limitations, we augment them with literature retrieval, which is essential for grounding identifying limitations in prior scientific findings.
Our approach enhances the capabilities of LLM systems to generate limitations in research papers, enabling them to provide more concrete and constructive feedback.

\begin{small}
\begin{center}
\begin{tabular}{cll}
\huggingface & \textbf{Data} & \href{https://huggingface.co/datasets/yale-nlp/LimitGen} {\path{yale-nlp/LimitGen}}\\
\github & \textbf{Code} & \href{https://github.com/yale-nlp/LimitGen}{\path{yale-nlp/LimitGen}}\\
\end{tabular}
\end{center} 
\end{small}
\end{abstract}

\begin{figure}[!t]
\centering
\includegraphics[width=1\linewidth]{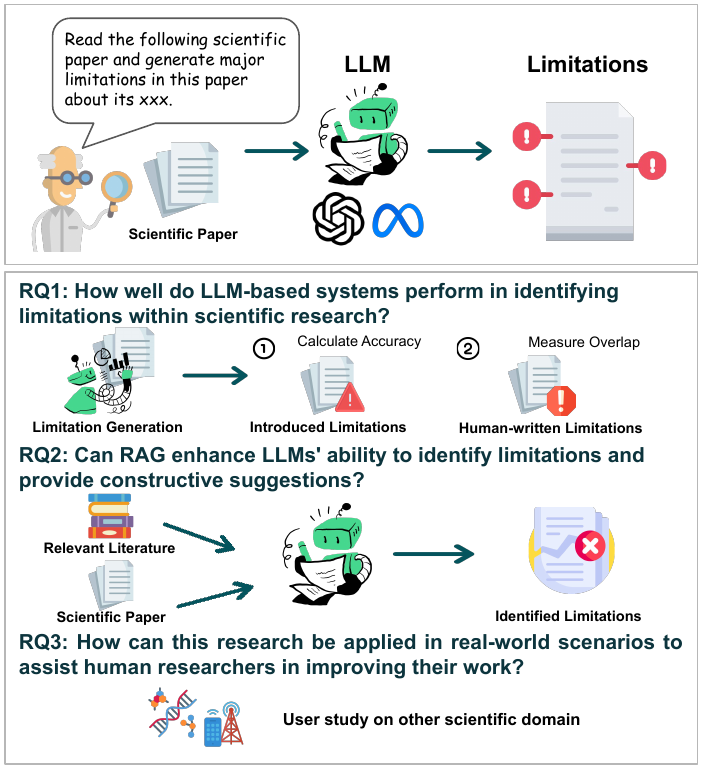}
\caption{Overview of the research: the limitation generation task and three research questions investigated.
}
\label{fig:pipeline}
\end{figure}
\section{Introduction}
Peer review plays a crucial role in ensuring the quality and integrity of scientific research. However, it is often a time-consuming and expertise-intensive process, posing significant challenges, especially as the volume of published papers continues to grow. Recent advancements in large language models (LLMs) have demonstrated remarkable capabilities across a variety of scientific tasks, such as answering questions about scientific papers~\cite{xu-etal-2024-kiwi}, writing scientific papers~\cite{chamoun-etal-2024-automated, lu2024ai}, retrieving related works~\cite{ajith-etal-2024-litsearch, press2024citeme}, improving idea generation~\cite{wang2024scimon, zhou2024hypothesis, si2024can}, and generating code to perform data-driven discovery~\cite{huang2024mlagentbench, tian2024scicode}. Meanwhile, there is increasing interest in exploring the potential of LLMs to assist with or generate peer reviews~\cite{liang2024can, liu2023reviewergpt, d2024marg, lou2024aaar}, which can be used to provide quick and early feedback to researchers and potentially alleviate some of the burdens associated with traditional review processes.

High-quality reviews are supposed to pinpoint the limitations of a paper and provide concrete, actionable suggestions, assisting researchers in improving their work. Existing benchmarks for peer-review generation collect papers and their corresponding entire reviews from AI conferences~\cite{du-etal-2024-llms, yu2024automated, tan2024peer}, but these benchmarks generally do not emphasize the importance of limitation identifications. Instead, they compare the overall quality of LLM-generated versus human-written reviews ~\cite{liang2024can, yu2024automated} and assess adherence to conference guidelines~\cite{tyser2024ai}. 
As opposed to other aspects of review generation, such as summary, strengths, syntactic or structural errors or request for elaboration, identification of limitations is the key aspect to facilitate future technical enhancement of  the work and it is of utmonst importance from the research growth point of view.

To address this gap, we present the first in-depth study and evaluation of LLM systems in identifying the limitations of scientific papers. To do so, we first provide a comprehensive taxonomy of types of limitations in scientific fields of study, with focus on AI.\footnote{We chose AI as this is the field we are familiar with. In \S\ref{sec:user-study} we also perform a user study to how our findings generalize to other domains.}
Then guided by this taxonomy, we propose \oursone, a synthetic benchmark focusing on various categories of limitations. \oursone systematically introduces controlled perturbations to high-quality papers to create scenarios where specific limitations are present. These perturbations include selective removal of crucial information such as experimental details, inadequate evaluation metrics, omission of key baseline comparisons, and constraints on datasets or methodologies. By carefully controlling these modifications, we can reliably evaluate how well LLM agents detect different types of limitations. In addition, \oursone also allows capturing suggestions on how to resolve the identified limitation.

To assess whether our taxonomy and the synthetic benchmark can effectively capture the diverse categories of limitations identified by humans in real-world peer review settings, we then collect human-written limitations from ICLR 2025 submissions as \ourstwo. We chose ICLR 2025 to mitigate contamination and also because ICLR peer reviews are often of high-quality due to the public nature of reviews and an extensive rebuttal process. Together, these two datasets form a comprehensive benchmark for advancing identification of limitations in papers. 

Limitation identification is a knowledge-intensive task, requiring years of expertise and staying current with rapidly evolving literature. In terms of modeling, in such settings, retrieval plays a crucial role as it is challenging to keep LLM agents up-to-date in rapidly evolving fields. While Retrieval-augmented generation (RAG) has been applied to enhancing scientific workflow, such as conducting literature reviews~\cite{agarwal2024litllm, asai2024openscholar} and answering domain-specific questions~\cite{xu-etal-2024-kiwi, skarlinski2024language}, it has not yet been explored in the context of peer-reviews. RAG also simulates how human identify limitations of papers by implicitly or explicitly referring to existing body of (often recent) literature and thus facilitate in grounding the generated limitations in existing scientific findings.
We enhance limitation generation by leveraging RAG techniques. Specifically, we prompt LLMs to query the Semantic Scholar API to retrieve papers related to the one under review, extracting relevant content to enrich their domain understanding. Our results demonstrate that incorporating RAG improves the ability of LLM systems to generate limitations in research papers, providing more concrete feedback.

We summarize our contributions as follows:
\begin{itemize} [leftmargin=*]
\itemsep0em 
\item We propose \ours, a comprehensive benchmark specifically designed to assess the ability of models to identify and address limitations in scientific research, with a reliable and systematic evaluation framework.

\item We evaluate the performance of LLMs and agent-based systems in identifying limitations and demonstrate their shortcomings in providing constructive and actionable feedback.

\item We explore the potential of RAG in review generation, demonstrating its ability to improve limitation identification and generate more contextually relevant and actionable suggestions.

\end{itemize}
\section{Related Work}
\subsection{Peer-Review Generation}

Recent advances in LLMs have significantly influenced scientific research, offering tools to streamline and enhance researchers' workflows across various stages of the scientific pipeline~\cite{xu-etal-2024-kiwi, lu2024ai, ajith-etal-2024-litsearch, wang2024scimon, zhou2024hypothesis, si2024can,  tian2024scicode}. 
Researchers have also extensively explored the potential of LLMs in automated peer review generation, employing various approaches such as guiding LLMs with single prompts~\cite{liang2024can}, adopting two-stage review generation frameworks with question-guided prompts~\cite{gao2024reviewer2}, and leveraging multi-agent systems~\cite{d2024marg}. Other studies simulate the complete review process as a multi-round dialogue~\cite{tan2024peer}. However, some research has found that LLM-generated reviews often suffer from generic and paper-unspecific content~\cite{du-etal-2024-llms}, are seldom entirely accurate, lack critical analysis, and fail to provide technical details~\cite{Zhou2024IsLA}. During the peer review process, identifying limitations is a crucial task as it helps highlight weaknesses in a study, guiding authors toward improvements and fostering scientific progress. 
Current research primarily focuses on generating the entire review. \citet{du-etal-2024-llms} collected human and LLM-generated reviews, each annotated by experts with fine-grained deficiency labels and explanations. \citet{tyser2024ai} compares generated reviews of papers with and without inserted errors by evaluating review scores. Other studies have constructed several exceptionally short computer science papers, each with an inserted error~\cite{liu2023reviewergpt}, or focused on identifying weaknesses within a single paragraph rather than an entire paper~\cite{chamoun-etal-2024-automated}.  \citet{lou2024aaar} extracted human-written weaknesses from peer reviews. However, these studies do not thoroughly evaluate whether LLM systems can effectively detect specific limitations in scientific research. In this work, we present a comprehensive benchmark to evaluate models' ability to identify and address limitations in AI research papers, comprising a synthetic subset created via controlled perturbations and a set of human-written limitations.

\subsection{Retrieval Augmented Generation}
Despite showing promise in various tasks, LLMs face significant challenges when adopted to specialized domains, including hallucinations~\cite{mallen-etal-2023-trust, mishra2024finegrained}, conflict between outdated pre-training data and latest domain knowledge~\cite{kasai2024realtime}, and lack of transparent attribution~\cite{ye-etal-2024-effective}.
Retrieval augmented generation that integrates external knowledge has emerged as a pivotal strategy to address these limitations~\cite{lewis2020retrieval, shuster-etal-2021-retrieval-augmentation, izacard2023atlas}, enabling LLMs to produce more accurate and context-aware outputs. 
 
Recent studies use proprietary LLMs with external APIs (\eg Semantic Scholar API \& Google Search API)~\cite{agarwal2024litllm, skarlinski2024language, chamoun-etal-2024-automated} or develop new methodologies to train specialized open models~\cite{asai2024openscholar} for tasks such as scientific literature review.
Furthermore, multiple-round retrieval-enhanced reasoning methods have been developed to improve retrieval effectiveness~\cite{he2022rethinking, shao-etal-2023-enhancing, jiang-etal-2023-active, chen-etal-2024-generalizing}. 
In this work, we introduce a novel approach that incorporates literature retrieval into the limitation generation process, enabling LLMs to utilize domain knowledge and produce more constructive feedback.

\section{\ours Benchmark}
This section discusses the task formulation of \ours and details the data construction process used to curate its two subsets.

\subsection{Task Formulation}
\begin{table*}[!t]
\centering
\small
\renewcommand{\arraystretch}{1.0}
\begin{tabular}{p{0.16\textwidth}p{0.2\textwidth}p{0.54\textwidth}}
\toprule
\textbf{Aspect} & \textbf{Limitation Subtype} & \textbf{Definition and Corresponding Data Example} \\
\midrule

\multirow{5}{*}{Methodology}  & Low Data Quality & The data collection method is unreliable, potentially introducing bias and lacking adequate preprocessing \hfill (\autoref{fig:perturb_type_1}) \\
\noalign{\vskip 0.5ex}\cdashline{2-3}[.4pt/1pt]\noalign{\vskip 0.5ex}
& Inappropriate Method & Some methods in the paper are unsuitable for addressing this research question and may lead to errors or oversimplifications \hfill (\autoref{fig:perturb_type_2}) \\

\midrule
\multirow{9}{*}{Experimental Design} &  Insufficient Baselines & Fail to evaluate the proposed approach against a broad range of well-established methods \hfill (\autoref{fig:perturb_type_3})\\

\noalign{\vskip 0.5ex}\cdashline{2-3}[.4pt/1pt]\noalign{\vskip 0.5ex}

& Limited Datasets & Rely on limited datasets, which may hinder the generalizability and robustness of the proposed approach  \hfill (\autoref{fig:perturb_type_4}) \\

\noalign{\vskip 0.5ex}\cdashline{2-3}[.4pt/1pt]\noalign{\vskip 0.5ex}
& Inappropriate Datasets & Use of inappropriate datasets, which may not accurately reflect the target task or real-world scenarios \hfill (\autoref{fig:perturb_type_5})\\

\noalign{\vskip 0.5ex}\cdashline{2-3}[.4pt/1pt]\noalign{\vskip 0.5ex}
& Lack of Ablation Studies & Fail to perform an ablation study, leaving the contribution of a certain component to the model's performance unclear \hfill (\autoref{fig:perturb_type_6})\\
\midrule
\multirow{5}{*}{Result Analysis} &  Limited Analysis & Rely on insufficient evaluation metrics, which may provide an incomplete assessment of the model's overall performance \hfill (\autoref{fig:perturb_type_7})\\

\noalign{\vskip 0.5ex}\cdashline{2-3}[.4pt/1pt]\noalign{\vskip 0.5ex}

& Insufficient Metrics & Offer insufficient insights into the model's behavior and failure cases  \hfill (\autoref{fig:perturb_type_8}) \\

\midrule
\multirow{9}{*}{Literature Review} &  Limited Scope & The review may focus on a very specific subset of literature or methods, leaving out important studies or novel perspectives \hfill (\autoref{fig:perturb_type_9})\\

\noalign{\vskip 0.5ex}\cdashline{2-3}[.4pt/1pt]\noalign{\vskip 0.5ex}

& Irrelevant Citations & Include irrelevant references or outdated methods, which distract from the main points and undermine the strength of conclusions  \hfill (\autoref{fig:perturb_type_10}) \\

\noalign{\vskip 0.5ex}\cdashline{2-3}[.4pt/1pt]\noalign{\vskip 0.5ex}
& Inaccurate Description & Provide an inaccurate description of existing methods, which can hinder readers' understanding of the context and relevance of the proposed approach \hfill (\autoref{fig:perturb_type_11})\\
\bottomrule
\end{tabular}

\caption{
The types of scientific paper limitations included in the \oursone subset.}
\label{tab:perturb}
\end{table*}
 We formally define the task of limitation generation in the context of LLMs as follows:
 Given: (1) a scientific paper, which may either contain a major limitation explicitly introduced (\ie \oursone discussed in \S\ref{sec:task1}), or exhibit limitations previously identified during peer review by human reviewers (\ie \ourstwo discussed in \S\ref{sec:task2}); and (2) an aspect of limitations, which serves as a focus point for the LLM to evaluate a specific dimension of the paper's quality. The LLM is tasked with generating the limitation for the given paper, reflecting its quality with respect to the specified aspect.
 
\subsection{Desiderata and Taxonomy of Limitations}
The identification and categorization of limitations in scientific research require careful consideration of what constitutes a meaningful limitation. Through our pilot analysis of peer reviews, we establish several key desiderata that guide our taxonomy of limitations.
First, a research limitation should represent a substantive constraint or weakness that impacts the validity, generalizability, or reliability of the study's findings. These constraints may arise from methodological choices, resource limitations, or gaps in current scientific understanding. Importantly, limitations should be distinguished from superficial critiques of presentation style.
Second, limitations should be actionable - they should point to specific aspects of the research that could be improved through concrete steps. This ensures that identifying limitations serves a constructive purpose in advancing scientific knowledge, rather than simply highlighting unavoidable constraints. For instance, a limitation regarding insufficient experimental validation should suggest specific additional experiments that would strengthen the work.
Third, limitations should be grounded in established scientific principles and practices within the relevant domain. This requires domain expertise to properly identify and articulate limitations that reflect meaningful departures from best practices or gaps in scientific rigor. For instance, appropriate evaluation metrics for each task are well-known within each subfield. 
Based on these desiderata and our analysis of peer review comments from top AI conferences, we categorize research limitations into four primary aspects (Table \ref{tab:perturb}):
\textit{(i) Methodological Limitations} focus on the fundamental approaches and techniques employed in the research. These include issues such as inappropriate choice of methods, unstated assumptions that may not hold, and problems with data quality or preprocessing that could introduce bias. Such limitations directly impact the validity of the research findings.
\textit{(ii) Experimental Design Limitations} encompass weaknesses in how the research validates its claims. This category includes insufficient baseline comparisons, limited datasets that may not represent the full problem space, and lack of ablation studies to isolate the contribution of different components. These limitations affect the reliability and reproducibility of results.
\textit{(iii) Results and Analysis Limitations} relate to how findings are evaluated and interpreted. This includes using inadequate evaluation metrics that may not capture important aspects of performance, insufficient error analysis, and lack of statistical significance testing. These limitations impact the strength and generalizability of conclusions.
\textit{(iv) Literature Related Limitations} focus on how the research connects to and builds upon existing work. This includes missing citations of relevant prior work, mischaracterization of existing methods, and failure to properly contextualize contributions within the broader research landscape. These limitations affect both the novelty claims and the proper attribution of ideas.
This taxonomy guides our creation of the \ours benchmark by ensuring we systematically evaluate different types of limitations that matter for scientific rigor. For each aspect, we identify specific subtypes of limitations that commonly appear in peer reviews and can be reliably assessed. The taxonomy also informs our evaluation criteria, as different types of limitations may require different forms of evidence and levels of domain knowledge to properly identify.

\subsection{\oursone Subset Collection}\label{sec:task1}
\paragraph{Source Paper Collection}
We collect scientific papers from arXiv under the ``Computation and Language'' category, focusing on those released between March 1, 2024, and May 31, 2024, a period likely outside the pretraining data cut-off for most current LLMs. This selection helps minimize potential data memorization issues that affect model evaluation. To extract content, we use the tool\footnote{\url{https://github.com/allenai/s2orc-doc2json}} by \citet{lo-etal-2020-s2orc}, which converts LaTeX source files into JSON format, capturing elements including the title, abstract, main sections, and appendix of each paper. In total, we compile an initial pool of 1,408 NLP papers for further annotation. We exclude papers that do not focus on experimental work, such as surveys, position papers, and dissertations, as these lack the experimental designs required for our analysis. Additionally, papers of insufficient quality are omitted to ensure that the introduced limitation represents the most critical issue in each paper. This filtering process led us to 500 papers.

\paragraph{Example Curation}

Following the taxonomy in \autoref{tab:perturb}, we design perturbation pipelines for each limitation subtype. For each paper, human experts determine the applicable perturbations and then apply all suitable perturbations accordingly. The annotators identify all the relevant sections in the paper based on the perturbation type. For each selected section, we employ GPT-4o to perturb the content according to the specific definitions and guidelines, such as removing relevant details or replacing a particular dataset. The prompts are provided in \autoref{fig:perturb_type_1} to \autoref{fig:perturb_type_11}.
Alongside each perturbation, we generate a brief description of the introduced limitation as the ground truth, which will serve as a reference for later evaluations.

\paragraph{Human Expert Validation}
To guarantee the reliability of our \oursone dataset, each annotated example is evaluated by a human annotator based on the following criteria: (1) The text within the paper must be grammatically correct and maintain clarity. (2) The introduced limitation must genuinely impact the quality and represent the most critical issue in the given aspect. (3) The generated ground truth limitation should clearly articulate the problem and be reasonable. Validators are tasked with revising or removing examples that do not meet these standards. In practice, from 500 papers, a total of 1,000 examples were retained, including 112 that were revised by human annotators. We provide the details of annotators involved in dataset construction in \autoref{tab:candidate_profiles}.

\begin{table}[!t]
\centering
\small
\addtolength{\tabcolsep}{-0.4em}
\renewcommand{\arraystretch}{1.1}
\begin{tabular}{lr}
\toprule
\textbf{Property} (\texttt{avg./max}) & \textbf{Value} \\
\midrule
\textbf{\oursone}\\
\quad Scientific Paper Word Length  & 5,201.46 / 58,788 \\
\quad Limitation Word Length  & 34.45 / 81 \\
\quad Paper Number & 500 \\
\quad Example Number & 1,000 \\

\midrule
\textbf{\ourstwo}\\
\quad Scientific Paper Word Length  & 8,255.38 / 1,8910 \\
\quad Limitation Word Length  & 61.97 / 795 \\
\quad Number of Limitations per Paper  & 6.05 / 20 \\
\quad Paper Number & 1,000 \\

\bottomrule
\end{tabular}
\caption{Data statistics of the \ours benchmark.}
\label{tab:stat}
\end{table}

\begin{figure}[!t]
\centering
\includegraphics[width=0.75\linewidth]{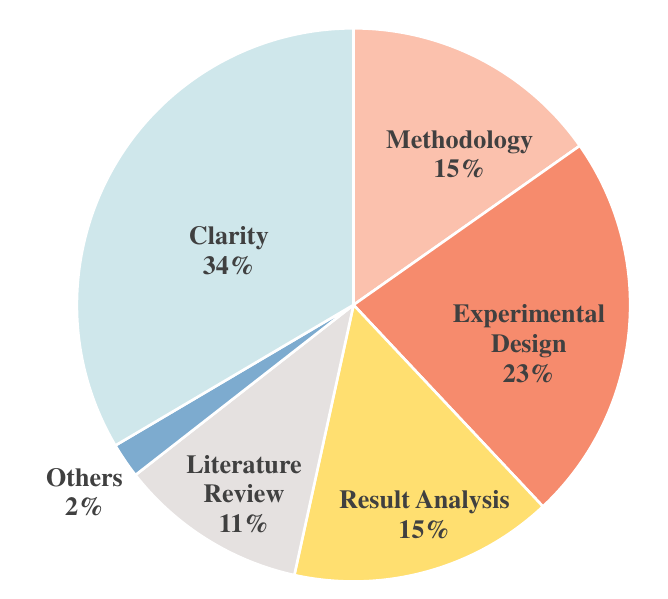}
\caption{The aspect distribution of human-written limitations in \ourstwo.
}
\label{fig:human_dist}
\end{figure}
\subsection{\ourstwo Subset Collection}\label{sec:task2}
To assess whether our taxonomy and the synthetic benchmark can effectively capture the diverse categories of limitations identified by humans in real-world peer review settings, we then collect human-written limitations from ICLR 2025 submissions.

We specifically focus on the weaknesses sections of each paper's reviews and break them down into itemized limitations.

To ensure quality, we use GPT-4o to exclude weaknesses that are too short (fewer than 20 words) or lack substantive suggestions, and then categorize the remaining limitations. The prompt is provided in \autoref{fig:prompt_filter}.

We retained only the limitations related to methodology, experimental design, result analysis, and literature review, considering them as ground truth. We collect a total of 9,844 papers and randomly sample 1,000 of them for experimentation.

\subsection{Data Statistics}
\autoref{tab:stat} illustrates the data statistics of our benchmark. \autoref{fig:human_dist} presents the detailed aspect distributions of the \oursone subset.
The full \ours benchmark consists of 2,000 examples and encompasses a diverse range of aspect types commonly found in paper limitations.

\section{\ours Evaluation Protocol}
Evaluating the quality of limitations generated by LLMs is inherently challenging due to the subjective and nuanced nature of research critique. Such assessments typically require expert-level judgment, making human evaluation labor-intensive. Moreover, comparing generated limitations with ground-truth ones is non-trivial,  as valid limitations may differ in phrasing or granularity.  These challenges motivate the careful design of our evaluation protocol to ensure both reliability and scalability.

\subsection{Human Evaluation Protocol}

\paragraph{Human Evaluation Process.}
For \oursone, we assess whether they correctly identify the intended subtype and calculate the accuracy. 
For \ourstwo, we assess the generated limitations across three dimensions: faithfulness, soundness, and importance. The detailed criteria are presented in the Appendix~\ref{app:guideline}.

For each criterion, Likert-scale scores ranging from 1 to 5 are used. Given the paper and a limitation generated by LLM, human evaluators are asked to assign scores for each dimension. Initially, ground truth references are not provided, minimizing potential bias from direct comparisons to the reference, as LLMs can generate limitations that are reasonable but not explicitly included in peer reviews. 

After submitting their initial scores, evaluators are then provided with the reference and asked to adjust their scores if they identify any aspects that may have been overlooked. 

\paragraph{Ensuring Reliable and Reproducible Human Evaluation.}
To ensure the reliability and reproducibility of our human evaluation, we develop a detailed assessment guideline, provided in Appendix~\ref{app:guideline}. To measure inter-annotator agreement, we sample 50 fixed generated instances from \oursone and \ourstwo, each independently assessed by two expert annotators. In \oursone, the resulting Cohen's Kappa score is 0.833. In \ourstwo, the scores for the criteria of importance, faithfulness, and soundness are 0.772, 0.735, and 0.717, respectively, indicating a high level of consistency among evaluators.

\begin{figure*}[!t]
\centering
\includegraphics[width=1\textwidth]{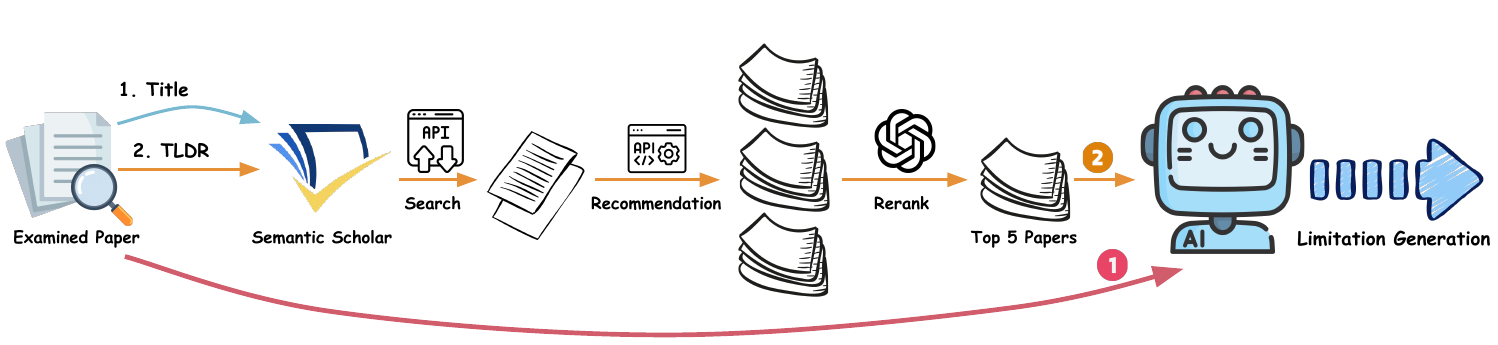}
\caption{An overview of RAG pipeline. We prompt LLMs to query the Semantic Scholar API, retrieve recommended papers, and rerank them based on their abstracts.
}
\label{fig:pipeline}
\end{figure*}
\subsection{Automated Evaluation Protocol}
To automatically evaluate the quality of the generated limitations, we compare them with the ground-truth limitations using a two-step process. 

\paragraph{Coarse-grained Evaluation.}
For \oursone, we use GPT-4o to classify the generated limitations and assess whether they correctly identify the intended subtype. Accuracy is used as the evaluation metric: a sample is deemed correct in the coarse-grained evaluation if at least one generated limitation accurately matches the subtype.
For \ourstwo, we refer to MARG~\cite{d2024marg}, evaluating recall, precision, and Jaccard Index to measure the overlap between generated and ground truth limitations for a paper. These metrics are then averaged across all papers to produce a single aggregated value for each metric.

\paragraph{Fine-grained Evaluation.} If a generated limitation correctly identifies the subtype or has a successful match in the ground truth limitations, we further evaluate the content to determine its alignment with the ground truth. This is achieved through reference-based evaluation using GPT-4o, which assigns scores to the generated limitations on from 1 to 5. These scores are based on two key criteria: relatedness to the ground truth and specificity in addressing the identified issue. Limitations that fail to determine the subtype or do not have a match during the coarse-grained evaluation are excluded from fine-grained evaluation and assigned a score of 0. For \oursone, we calculate the average of the highest scores assigned to the limitations of each paper in fine-grained evaluation. For \ourstwo, we calculate the average of all limitations for each paper and then compute the overall average across all papers. This provides a holistic measure of the system's performance across both accuracy and quality dimensions.

\paragraph{Reliability Assessment.}
To validate the performance of our automated evaluation system, we also calculate the system correlation between the automated fine-grained evaluation and the human evaluation, using data presented in \autoref{tab:results_1} and \autoref{tab:results_2}. 
In \oursone, the correlation between the fine-grained score and accuracy is 0.96. In \ourstwo, the correlation between the fine-grained score and faithfulness, soundness, and importance scores are 0.77, 0.60, and 0.67. By comparing with ground truth, our automated evaluation system can effectively assess the quality of the generated limitations.

\section{Evaluated Systems}
We next discuss the systems evaluated in our experiments, including LLMs, agent-based system, and RAG-enhanced pipeline.

\subsection{Evaluated LLMs}
We evaluate the performance of 4 frontier LLMs across two distinct categories in our benchmark: (1) Proprietary LLMs, including GPT-4o and GPT-4o-mini~\cite{openai2024gpt4o}; and (2) Open-source LLMs, including Llama-3.3-70B~\cite{dubey2024llama3}, Qwen2.5-72B~\cite{qwen2}.  
We require each model to generate the most significant limitations for an aspect of a paper. In the \oursone experiments, we measure whether models identify the single most prominent limitation in each paper within their top three generated limitations, ensuring a fair comparison across systems.

\subsection{Evaluated Agent-based System.}
We also present our multi-agent approach for generating limitations. Our architecture, following MARG \cite{d2024marg}, consists of a set of chat-based LLM agents (GPT-4o-mini in this study), each with its own chat history and prompt(s). The system includes three distinct agent roles: (1) a leader agent, responsible for coordinating tasks among agents; (2) a worker agent, which processes the full text of the paper; and (3) an expert agent, prompted to focus on a specialized sub-task to support the leader. With task instructions for each aspect, the leader delegates specific sub-tasks to the other agents and synthesizes their responses to produce the final limitations.

\subsection{RAG-Enhanced Limitation Generation}
In preliminary testing, we observed that LLMs often fail to detect limitations or provide substantive suggestions due to a lack of knowledge in related areas. To address this, we enhanced the evaluated systems' capabilities by incorporating the RAG module, a method proven effective for knowledge-intensive tasks \cite{lewis2020retrieval, shi-etal-2024-replug}, to ground limitation generation in the relevant literature. This method enables the LLMs to retrieve and consider related works when evaluating limitations in the given research paper.

Specifically, the retrieval process leverages the Semantic Scholar API and adapts based on the input paper's availability in the database. If the paper is available the database, we use its Semantic Scholar ID to fetch at most 20 recommended papers via the recommendation API\footnote{\url{https://api.semanticscholar.org/api-docs/recommendations}}. If the paper is unavailable, we use GPT-4o-mini to generate a query based on the paper's abstract and use the relevance API\footnote{\url{https://api.semanticscholar.org/api-docs/graph\#tag/Paper-Data/operation/get_graph_paper_relevance_search}} to identify related papers. From this search, the top 3 results are treated as seed papers, and for each seed paper, 5 additional recommendations are retrieved through the recommendation API, yielding a pool of 18 papers. These retrieved papers are then reranked by GPT-4o-mini, which assesses the similarity between the input paper and the candidates. The top 5 papers are selected. 

Due to LLMs' context window constraints, directly providing all retrieved papers to them is impractical. We employ GPT-4o-mini to identify and extract content related to methodology, experimental design, result analysis, and literature review. This extracted content is then concatenated and used as a concise reference to help the LLMs effectively identify limitations in these aspects. In experiments involving MARG, 
we enable the expert agent to retrieve related papers and provide specific suggestions based on the retrieved content while refining the initial limitation comments.
\section{Experiment Results}
This section presents our main findings and in-depth analysis.

\subsection{Results and Analysis}\label{sec:main-result}

\begin{tcolorbox}[colback=gray!10, colframe=white, width=\linewidth, boxrule=0.5mm, arc=1mm, outer arc=1mm, left=2mm, right=2mm]
\textcolor{cyan!80!black}{\faTools} \xspace 
\textbf{RQ1:} \rqone 
\end{tcolorbox}

\begin{table}[!t]
\centering
\resizebox{\linewidth}{!}{%
\small
\addtolength{\tabcolsep}{-0.1em}
\renewcommand{\arraystretch}{1.1}
\begin{tabular}{lccc}
\toprule
\multirow{2}{*}{\textbf{Systems}} & \multicolumn{2}{c}{\textbf{Automated Eval.}}  & \multicolumn{1}{c}{\textbf{Human Eval.}}\\
\cmidrule(lr){2-3} \cmidrule(lr){4-4}
& Coarse & Fine (0-5) & Accuracy \\
\midrule
Human & 86.0\% & 3.52 & 82.0\% \\
\midrule
GPT-4o & 52.0\% & 1.34  & 45.9\%\\\
\quad w/ RAG & +12.2\% & +0.37 & +16.0\% \\
\midrule
GPT-4o-mini & 49.1\% & 1.25 & 37.8\% \\
\quad  w/RAG & +4.2\% & +0.13 & +5.9\% \\
\midrule
Llama-3.3-70B & 45.7\% & 1.15 & 32.7\% \\
\quad  w/RAG & +2.4\% & +0.05 & +4.5\% \\
\midrule
Qwen-2.5-72B & 47.1\% & 1.20 & 31.5\% \\
\quad  w/RAG & +1.2\% & +0.03 & +3.9\% \\
\midrule
MARG & 68.1\% & 1.83 & 54.8\% \\
\quad  w/ RAG & +9.8\% & +0.27 & +17.7\% \\
\bottomrule
\end{tabular}
}

\caption{
Human and automated evaluation results of the LLMs and Agent-based system on \oursone set averaged across all subtypes. For human evaluation, we randomly sample 100 examples from the dataset.
}
\label{tab:results_1}
\end{table}

\noindent \autoref{tab:results_1} shows the performance of the evaluated systems on \oursone. In Appendix~\ref{app:exp}, we provide more detailed results on their performance for each subtype of limitation. The results demonstrate that identifying limitations in scientific papers remains a significant challenge for current LLMs. Even the best-performing LLM, GPT-4o, can only identify about half of the limitations that humans consider very obvious. Although MARG leverages multi-agent collaboration and generates more comments, successfully identifying more limitations, the feedback it provides still lacks specificity, which is reflected in the fine-grained scores.

\begin{table}[!t]
\centering
\resizebox{\linewidth}{!}{%
\small
\addtolength{\tabcolsep}{-0.3em}
\renewcommand{\arraystretch}{1.1}
\begin{tabular}{lccccc}
\toprule
\multirow{2}{*}{\textbf{Systems}} & \multicolumn{2}{c}{\textbf{Automated Eval.}}  & \multicolumn{3}{c}{\textbf{Human Eval. (1-5)}}\\
\cmidrule(lr){2-3} \cmidrule(lr){4-6}
& Jaccard & Fine.(0-5) & Faith. & Sound. & Import. \\
\midrule
GPT-4o & 15.9\% & 0.42 & 3.19 & 2.84 & 3.49 \\
\quad  w/ RAG & +2.9\% & +0.13 & +0.49 & +1.13 & +0.60 \\
\midrule
GPT-4o-mini & 15.5\% & 0.39 & 3.03 & 2.78 & 2.97 \\
\quad  w/ RAG & +0.6\% & +0.01 & +0.28 & +0.77 & +0.53 \\
\midrule
Llama-3.3-70B & 16.3\% & 0.39 & 2.98 & 2.85 & 3.05 \\
\quad  w/ RAG & +0.1\% & +0.04 & +0.23 & +0.70 & +0.21 \\
\midrule
Qwen-2.5-72B & 14.4\% & 0.53 & 2.91 & 2.86 & 2.94 \\
\quad  w/ RAG & +1.0\% & +0.11 & +0.22 & +0.35 & +0.34 \\
\midrule
MARG & 15.2\% & 0.66 & 3.60 & 3.19 & 3.78 \\
\quad  w/ RAG & +2.5\% & +0.24 & +0.52 & +0.98 & +0.43 \\

\bottomrule
\end{tabular}
}

\caption{
Human and automated evaluation results of the LLMs and Agent-based system on \ourstwo set averaged across all aspects. We randomly sample 100 examples from the dataset for human evaluation.
}
\label{tab:results_2}
\end{table}
\autoref{tab:results_2} shows the performance of the evaluated systems on \ourstwo, while the results on their performance for each aspect are illustrated in Appendix~\ref{app:exp}. MARG outperforms all LLMs in terms of fine-grained scores and human evaluation but generates more comments than the other baselines, resulting in lower Jaccard scores. Consistent with the findings on \oursone, the performance of all systems in \ourstwo remains quite poor. Their generated insights and feedback for top AI conference submissions lack depth and inspiration, especially when compared to those provided by experienced reviewers.

\subsection{Analysis of RAG Pipeline}
\begin{tcolorbox}[colback=gray!10, colframe=white, width=\linewidth, boxrule=0.5mm, arc=1mm, outer arc=1mm, left=2mm, right=2mm]
\textcolor{cyan!80!black}{\faTools} \xspace 
\textbf{RQ2:} \rqtwo
\end{tcolorbox}
\paragraph{Overall Results.} 
While LLMs currently struggle to identify limitations in scientific papers and provide constructive advice, there is potential for them to offer better feedback if they can retrieve relevant literature to address their gaps in domain knowledge and understanding of the research context. We conducted experiments on all evaluated systems with the integration of the RAG pipeline. As shown in \autoref{tab:results_1} and \autoref{tab:results_2}, incorporating the RAG method can enhance LLM performance in refining their outputs.

\paragraph{Impact of Retrieved Content Quality on LLM Performance.}
We also investigate the impact of the quality of retrieved content on LLM performance. In \ourstwo, we randomly sample 100 examples and conduct experiments by GPT-4o-mini. For each example, we provide another two sets of retrieved papers as references: the top 3 ranked papers and the last 5 papers after re-ranking, from the 18 retrieved papers.
The results, as shown in \autoref{tab:results_ablation}, demonstrate that providing a broader set of relevant papers, as in the standard RAG method with the top 5 papers, improves the LLM's performance in generating accurate limitations compared to using only the top 3 or the last 5 papers. RAG consistently provides some benefits, even when the retrieved papers are not the most relevant.
\begin{table}[!t]
\centering
\resizebox{\linewidth}{!}{%
\small
\addtolength{\tabcolsep}{-0.4em}
\renewcommand{\arraystretch}{1.1}
\begin{tabular}{lccccc}
\toprule
\multirow{2}{*}{\textbf{Systems}} & \multicolumn{2}{c}{\textbf{Automated Eval.}}  & \multicolumn{3}{c}{\textbf{Human Eval. (1-5)}}\\
\cmidrule(lr){2-3} \cmidrule(lr){4-6}
& Jaccard & Fine.(0-5) & Faith. & Sound. & Import. \\
\midrule
GPT-4o-mini & 15.0\% & 0.36 & 3.03 & 2.78 & 2.97 \\
\quad  w/ RAG (Top 5) & +1.4\% & +0.05 & +0.28 & +0.77 & +0.53 \\
\quad  w/ RAG (Top 3) & +1.3\% & +0.04 & +0.19 & +0.56 & +0.31 \\
\quad  w/ RAG (Last 5) & +0.8\% & +0.03 & +0.07 & +0.09 & +0.05 \\

\bottomrule
\end{tabular}
}

\caption{
Human and automated evaluation results of different RAG settings from 18 retrieved papers on the subset of 100 examples from \ourstwo.
}
\label{tab:results_ablation}
\end{table}

\paragraph{Case Study.}
We further conduct a case study to analyze the impact of RAG on LLM systems' ability to identify limitations. We select a total of 20 examples from both subsets, each successfully matching the targeted limitation subtype or receiving all three ratings of 4 or higher in human evaluation. See the Appendix~\ref{app:case} for some of the examples. Retrieved external knowledge provides LLMs with up-to-date domain information and offers standard practices for addressing specific issues. By comparing relevant papers with the examined paper, LLM systems are better equipped to identify problems. Systems with stronger reasoning capabilities, such as GPT-4o and MARG, benefit the most from RAG, as they can leverage external information to derive meaningful insights and improve their analysis.

\subsection{User Studies on Real-world Scenarios}\label{sec:user-study}

\begin{tcolorbox}[colback=gray!10, colframe=white, width=\linewidth, boxrule=0.5mm, arc=1mm, outer arc=1mm, left=2mm, right=2mm]
\textcolor{cyan!80!black}{\faTools} \xspace
\textbf{RQ3:} \rqthree
\end{tcolorbox}

\begin{table}[!t]
\centering
\small
\renewcommand{\arraystretch}{1.1}
\addtolength{\tabcolsep}{-0.1em}
\begin{tabular}{ll}
\toprule
\textbf{User Study} & \textbf{Acc.} \\
\midrule

\textbf{GPT-4o} \\
\quad NLP Domain (as \oursone) & 45.9\%\\
\quad Biomedical Domain & 31.3\%\\
\quad Computer Network Domain & 37.5\%\\
\textbf{GPT-4o w/ RAG} \\
\quad NLP Domain (as \oursone) & 61.9\% \\
\quad Biomedical Domain & 50.0\%\\
\quad Computer Network Domain & 56.3\%\\
\noalign{\vskip 0.5ex}\hdashline\noalign{\vskip 0.5ex}
\textbf{Llama-3.3-70B} \\
\quad NLP Domain (as \oursone) & 32.7\% \\
\quad Biomedical Domain & 25.0\% \\
\quad Computer Network Domain & 31.3\%\\
\textbf{Llama-3.3-70B w /RAG} \\
\quad NLP Domain (as \oursone) & 37.2\% \\
\quad Biomedical Domain & 31.3\%\\
\quad Computer Network Domain & 37.5\%\\

\bottomrule
\end{tabular}

\caption{Human evaluation result of the adaptability of our research across different scientific domains.}
\label{tab:user_study}
\end{table}

\noindent Our research focuses primarily on the AI domains. To investigate the applicability of our findings in more real-world scenarios, we design the following user studies to explore the domain generalization of our research. Specifically, we examine the areas of biomedical sciences and computer networks. We first engage two experts in the two domains, each providing five research papers from their respective fields, focusing on those published after May 15, 2024, with which they are familiar. Following the annotation procedure outlined in \oursone, the experts design perturbations across four aspects and annotate 32 examples in total. We then present another two experts with the perturbed papers and the generated limitations under two conditions: one utilizing our RAG pipeline and one without. As shown in \autoref{tab:user_study}, the human evaluation scores for GPT-4o and Llama-3.3-70B are consistent with the results observed in our main experiments. Our retrieval pipeline enhances the ability of LLMs to identify limitations. We believe that future work could further extend our research framework to encompass additional scientific domains.
\section{Conclusion}
This paper presents \ours, the first benchmark designed for systematically evaluating models on identifying and addressing scientific research limitations, supported by a reliable and systematic evaluation framework. We also demonstrate how RAG enhances limitation generation, showcasing its ability to help models identify weaknesses and provide more constructive feedback. Through a comprehensive analysis of LLM-based approaches for identifying different types of limitations, we offer key insights to guide future advancements.

\section*{Acknowledgments}
This project is supported by Tata Sons Private Limited, Tata Consultancy Services Limited, and Titan. 
We are grateful to Nvidia Academic Grant Program for providing computing resources.

\section*{Limitations}
While our study provides valuable insights into the ability of LLM systems to identify limitations in scientific papers, several limitations remain that present opportunities for future work.

First, our work does not include non-textual inputs such as figures, which are integral to many scientific papers. As figures often provide crucial evidence or highlight key findings, future extensions to our benchmark could incorporate multimodal inputs to better evaluate LLMs' ability to identify limitations arising from inconsistencies or omissions in visual data.

Second, this study does not explore advanced RAG techniques. Our focus is on assessing the potential of LLM systems in this context rather than optimizing retrieval methods. We encourage researchers to build upon our benchmark and investigate advanced retrieval methods to further improve limitation identification.

Lastly, while our benchmark offers valuable insights into model performance, there are several limitations that should be considered. The current benchmark covers a limited time span, including some parts of 2024 and ICLR 2025, which may not fully represent the evolving landscape of research in the field. Given the rapid advancements in NLP, it is important to regularly update the benchmark to incorporate the latest publications. Another potential limitation lies in the reliance on our automated evaluation method. Inherent biases in these systems could affect the accuracy and reliability of the overall evaluation. Additionally, our taxonomy and benchmark focus primarily on AI, as this is the field we are most familiar with. Although we conducted a user study to assess its applicability to other domains, the nuances of different scientific disciplines may introduce challenges and limitation types that our framework does not fully address. Future work could expand this taxonomy by collaborating with experts from diverse fields, such as medicine, physics, and social sciences, to ensure broader generalizability.

\section*{Ethical Considerations}
 We have carefully considered the ethical implications of our work, which focuses on identifying limitations in scientific papers. Our approach is designed to assist human reviewers by offering complementary insights rather than replacing their essential role in the peer review process. We acknowledge potential risks, such as biases in LLM-generated outputs and the potential to undermine the integrity of scientific evaluations if these systems are misused. Our study emphasizes that LLMs are far from achieving the level of expertise and nuanced understanding of human experts. Future developments in this field should prioritize transparency, fairness, and risk mitigation to ensure these tools are employed responsibly. 
Furthermore, the raw paper data used in our study is collected from arXiv with distributed under the CC BY 4.0 (Creative Commons Attribution 4.0 International) license. In alignment with this licensing framework, we will release our dataset under the same CC BY 4.0 license.
This ensures that our dataset remains freely accessible while requiring proper attribution to the original sources, thereby maintaining legal and ethical compliance with the terms under which the original data was shared.

\bibliography{anthology,custom,llm}

\appendix

\newpage
\section{\ours Benchmark}\label{app:data}
\begin{table*}[h]
\centering
\small
\begin{tabular}{llcccc}
\toprule
\textbf{ID} & \textbf{\# NLP/AI Publication} & \textbf{Data Annotation} & \textbf{Data Validation} & \textbf{Human Evaluation} & \textbf{Human Performance}\\
\midrule
1 & > 10 & \cmark &  & \cmark & \\

2 & > 10 & & \cmark &  & \cmark  \\

3 & 5-10 & \cmark & \cmark & \\

4 & 5-10 & \cmark &  & \cmark & \\

5 & 1-5 & \cmark & \\

6 & 1-5 & \cmark & & & \cmark \\
\bottomrule
\end{tabular}
\caption{Details of annotators involved in dataset construction and LLM performance evaluation. \ours is annotated by experts in NLP domains, ensuring both the accuracy of the benchmark and the reliability of the human evaluation.}
\label{tab:candidate_profiles}
\end{table*}
\subsection{\oursone}
\autoref{tab:stat_one} illustrates the detailed distribution of the introduced limitation subtypes in our \oursone.

\subsection{\ourstwo}
We randomly sample 1,000 papers from the ICLR 2025 submissions. We use GPT-4o to filter and classify the ground truth limitations, with the prompt provided in  \autoref{fig:prompt_filter} and \autoref{fig:prompt_class}.
\begin{table}[!t]
\centering
\small
\addtolength{\tabcolsep}{-0.4em}
\renewcommand{\arraystretch}{1.1}
\begin{tabular}{lr}
\toprule
\textbf{Property} & \textbf{Value} \\
\midrule
\textbf{Methodology} & 250\\
\quad \# Low Data Quality & 125 \\
\quad \# Inappropriate Method & 125 \\

\midrule
\textbf{Experimental Design} & 250\\
\quad \# Insufficient Baseline & 62 \\
\quad \# Limited Datasets & 63 \\
\quad \# Inappropriate Datasets & 63 \\
\quad \# Lack of Ablation Study & 62 \\

\midrule
\textbf{Result Analysis} & 250\\
\quad \# Limited Analysis & 125 \\
\quad \# Insufficient Metrics  & 125 \\

\midrule
\textbf{Experimental Design} & 250\\
\quad \# Limited Scope & 83 \\
\quad \# Irrelevant Citations & 84 \\
\quad \# Inaccurate Description & 83 \\

\bottomrule
\end{tabular}

\caption{Subtype distribution in the \oursone subset.}
\label{tab:stat_one}
\end{table}

\subsection{Annotator Guidelines}\label{app:guideline}

All annotators are experts with several NLP/ML publications as shown in \autoref{tab:candidate_profiles}. To ensure quality, they follow detailed annotation guidelines, which provide clear instructions for the annotation process.

\paragraph{Source Paper Collection}
Our annotators follow the guidelines below to ensure only well-written arXiv papers are selected for perturbation:

\begin{itemize}[leftmargin=*]
    \itemsep0em
    \item Exclude papers that do not focus on experimental work, such as surveys, position papers, and dissertations.
    \item Avoid papers with poorly written sections, lack of structure, or unprofessional presentation..
    \item Ensure the methods are well-defined, reproducible, and grounded in established scientific principles. Avoid papers with vague or unsupported claims.
    \item Select papers that provide thorough experiments, proper baselines, and detailed evaluations. The results should be well-documented and statistically sound.
    \item The paper should present a meaningful contribution to the field, such as a novel approach, insights, or applications, rather than incremental work.
\end{itemize}

\paragraph{Data Validation}
When validating the perturbation, annotators should follow these guidelines carefully to evaluate whether the perturbations meet the intended quality standards:

\begin{itemize}[leftmargin=*]
    \itemsep0em
    \item Check that the generated perturbation aligns with the limitation type specified in the instruction and verify that GPT-4o strictly follows the provided instruction to introduce the intended limitation.
    \item Ensure that all relevant sections needing modification are appropriately updated.
    \item Confirm that the perturbation does not compromise the clarity of the original text.
    \item Verify that the introduced limitation represents the most evident and significant limitation of the targeted aspect.
    \item Ensure the introduction of the limitation does not lead to unintended limitations elsewhere in the paper.
    \item The generated ground truth limitation should clearly articulate the problem and be reasonable.

\end{itemize}

\paragraph{Human Evaluation}
For \ourstwo, we assess the generated limitations across three dimensions:

\begin{itemize}[leftmargin=*]
    \itemsep0em
    \item \textbf{Faithfulness:} The generated limitations should accurately represent the paper's content and findings, avoiding any introduction of misinformation or contradictions to the original concepts, methodologies or results presented.
    \begin{itemize}[leftmargin=*]
        \itemsep0em
        \item 5 points: Perfect alignment with the original content and findings, with no misinformation or contradictions. Fully reflects the paper’s concepts, methodologies, and results accurately.
        \item 4 points: Mostly aligns with the original content but contains minor inaccuracies or slight misinterpretations. These do not significantly affect the overall understanding of the paper’s concepts or results.
        \item 3 points: Generally aligns with the original content but includes several minor inaccuracies or contradictions. Some elements may not fully reflect the paper’s concepts or results, though the overall understanding is mostly intact.
        \item 2 points: Noticeable misalignment with the original content, with multiple inaccuracies or contradictions that could mislead readers. Some key aspects of the paper’s concepts or results are misrepresented.
        \item 1 point: Introduces significant misalignment by misrepresenting issues that do not exist in the paper. Creates considerable misinformation and contradictions that distort the original content, concepts, or results.
    \end{itemize}
    \item \textbf{Soundness:} The generated limitations should be detailed and specific, with suggestions or critiques that are practical, logically coherent, and purposeful. It should clearly address relevant aspects of the paper and offer insights that can genuinely improve the research.
    \begin{itemize}[leftmargin=*]
        \itemsep0em
        \item 5 points: Highly detailed and specific, with practical, logically coherent, and purposeful suggestions. Clearly addresses relevant aspects and offers insights that substantially improve the research.
        \item 4 points: Detailed and mostly specific, with generally practical and logically sound suggestions. Addresses relevant aspects well but may lack depth or novelty in some areas.
        \item 3 points: Detailed and specific but with some issues in practicality or logical coherence. Suggestions are somewhat relevant and offer partial improvements.
        \item 2 points: Somewhat vague or lacking in specificity, with suggestions that have limited practicality or logical coherence. Addresses relevant aspects only partially and provides minimal improvement.
        \item 1 point: Lacks detail and specificity, with impractical or incoherent suggestions. Fails to effectively address relevant aspects or offer constructive insights for improvement.

    \end{itemize}
    \item \textbf{Importance:} The generated limitations should address the most significant issues that impact the paper's main findings and contributions. They should highlight key areas where improvements or further research are needed, emphasizing their potential to enhance the research's relevance and overall impact.
    \begin{itemize}[leftmargin=*]
        \itemsep0em
        \item 5 points: Addresses critical issues that substantially impact the paper’s findings and contributions. Clearly identifies major areas for significant improvement or further research, enhancing the research’s relevance and overall impact.
        \item 4 points: Identifies meaningful issues that contribute to refining the paper’s findings and methodology. While the impact is notable, it does not reach the level of fundamentally shaping future research directions.
        \item 3 points: Highlights important issues that offer some improvement to the current work but do not significantly impact future research directions. Provides useful insights for refining the paper but lacks broader implications for further study.
        \item 2 points: Points out limitations with limited relevance to the paper’s overall findings and contributions. Suggestions offer marginal improvements but fail to address more substantial gaps in the research.
        \item 1 point: Focuses on trivial issues, such as minor errors or overly detailed aspects. Does not address substantive issues affecting the paper’s findings or contributions, limiting its overall relevance and impact.

    \end{itemize}
\end{itemize}

\subsection{Human Baseline}\label{app:human-performance}
To obtain an informative estimate of expert-level performance on \ours, we randomly sample 50 examples from each subset. 
Two expert annotators (\ie Annotators 1 and 6, as described in \autoref{tab:candidate_profiles}) independently solve these examples.

During human evaluation, the expert evaluators are not informed of the sources of these generated limitations. We report the evaluation results on \autoref{tab:results_1} and \autoref{tab:results_2}.

\subsection{Limitation Taxonomy}
\begin{figure*}[!t]
\centering
\includegraphics[width=1\textwidth]{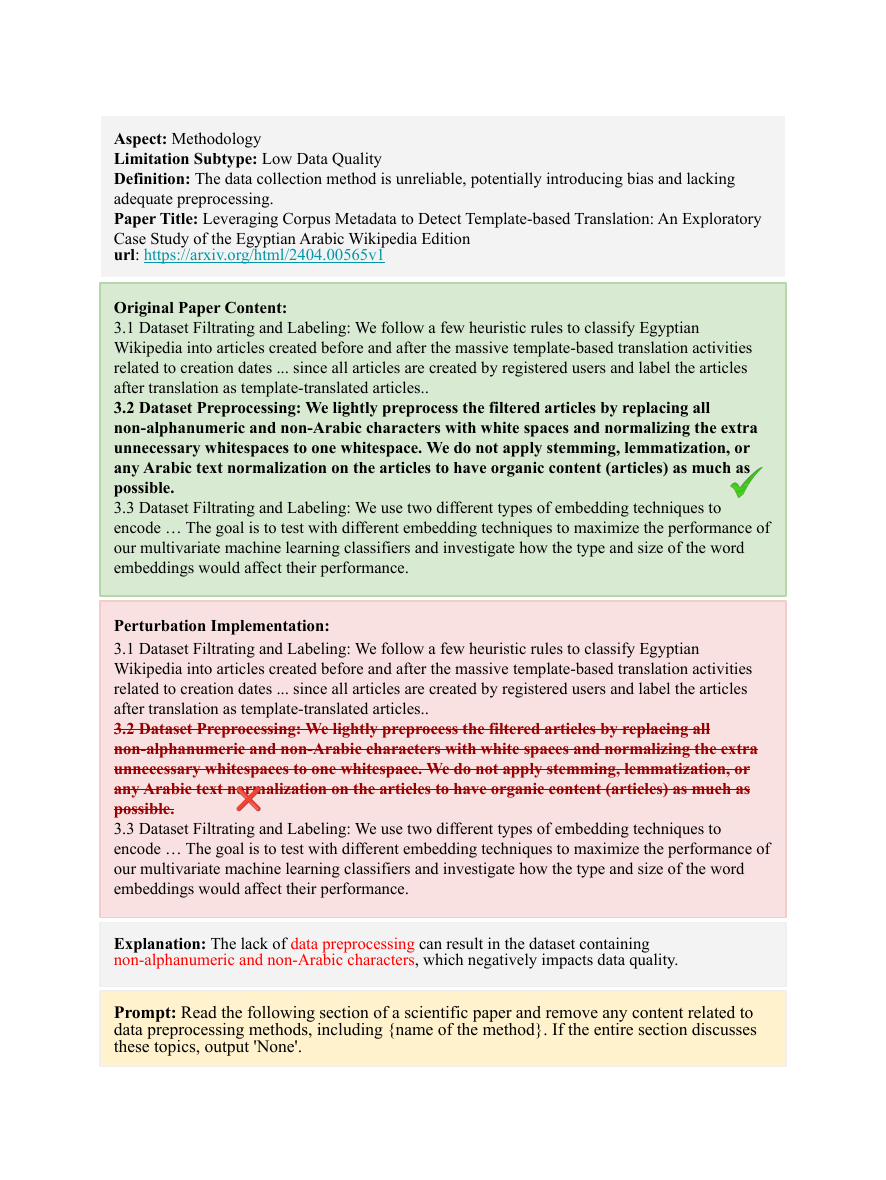}
\caption{
An example of \textbf{Low Data Quality} and its perturbation implementation.
}
\label{fig:perturb_type_1}
\end{figure*}

\begin{figure*}[!t]
\centering
\includegraphics[width=1\textwidth]{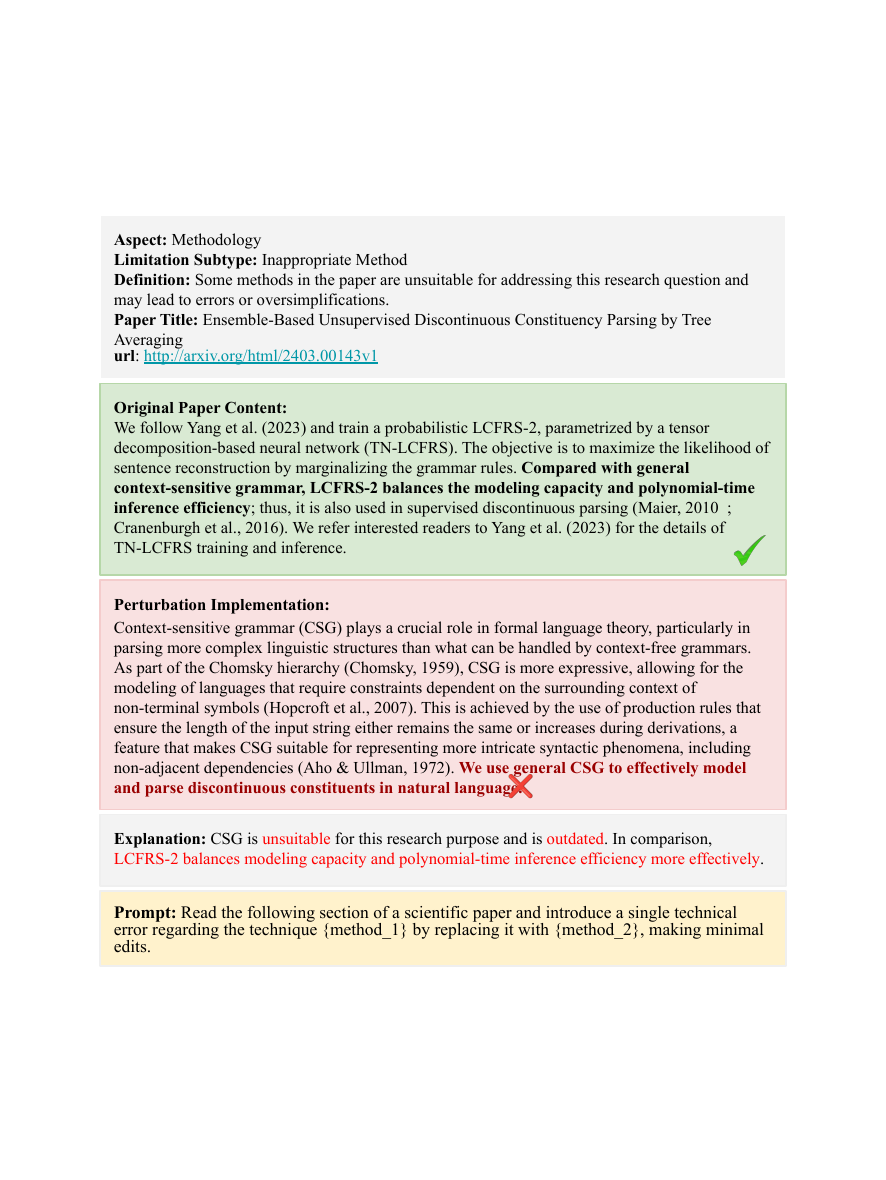}
\caption{
An example of \textbf{Inappropriate Method} and its perturbation implementation.
}
\label{fig:perturb_type_2}
\end{figure*}

\begin{figure*}[!t]
\centering
\includegraphics[width=1\textwidth]{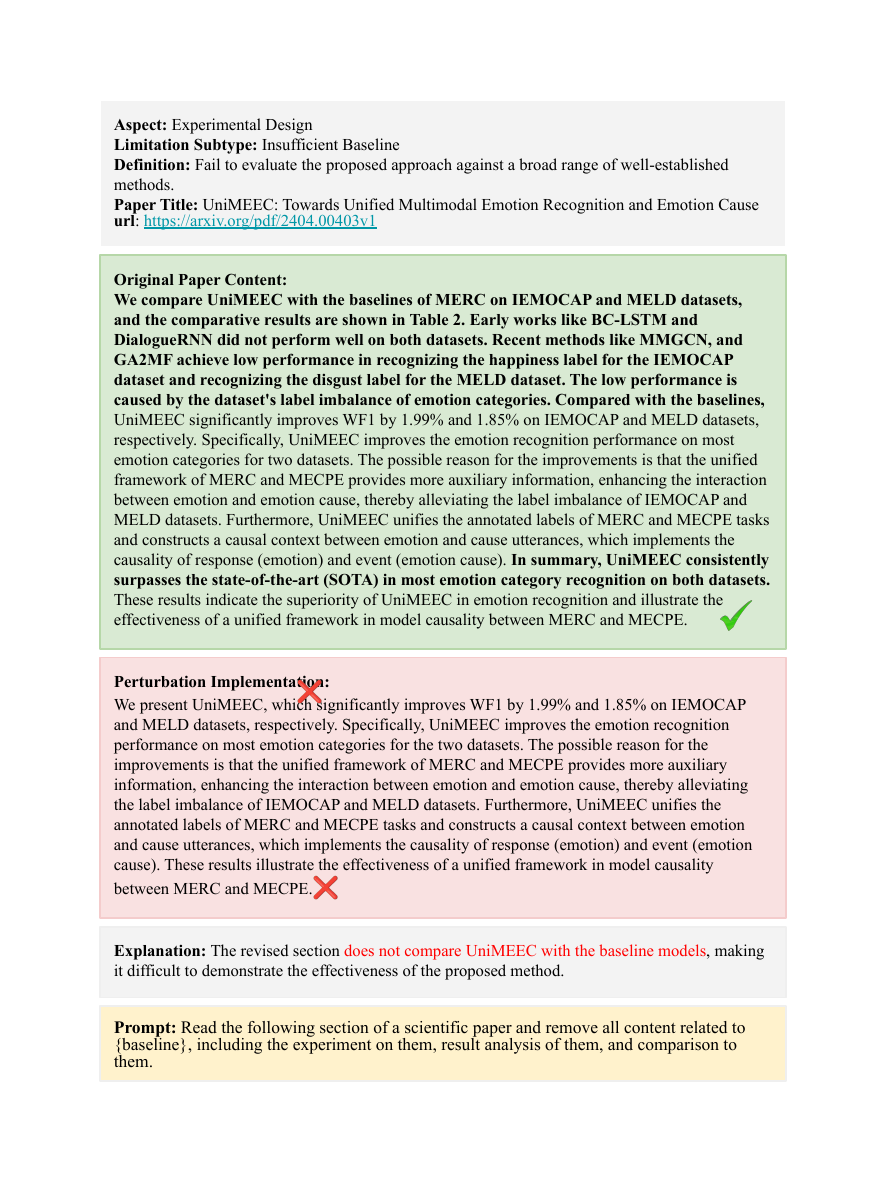}
\caption{
An example of \textbf{Insufficient Baseline} and its perturbation implementation.
}
\label{fig:perturb_type_3}
\end{figure*}

\begin{figure*}[!t]
\centering
\includegraphics[width=1\textwidth]{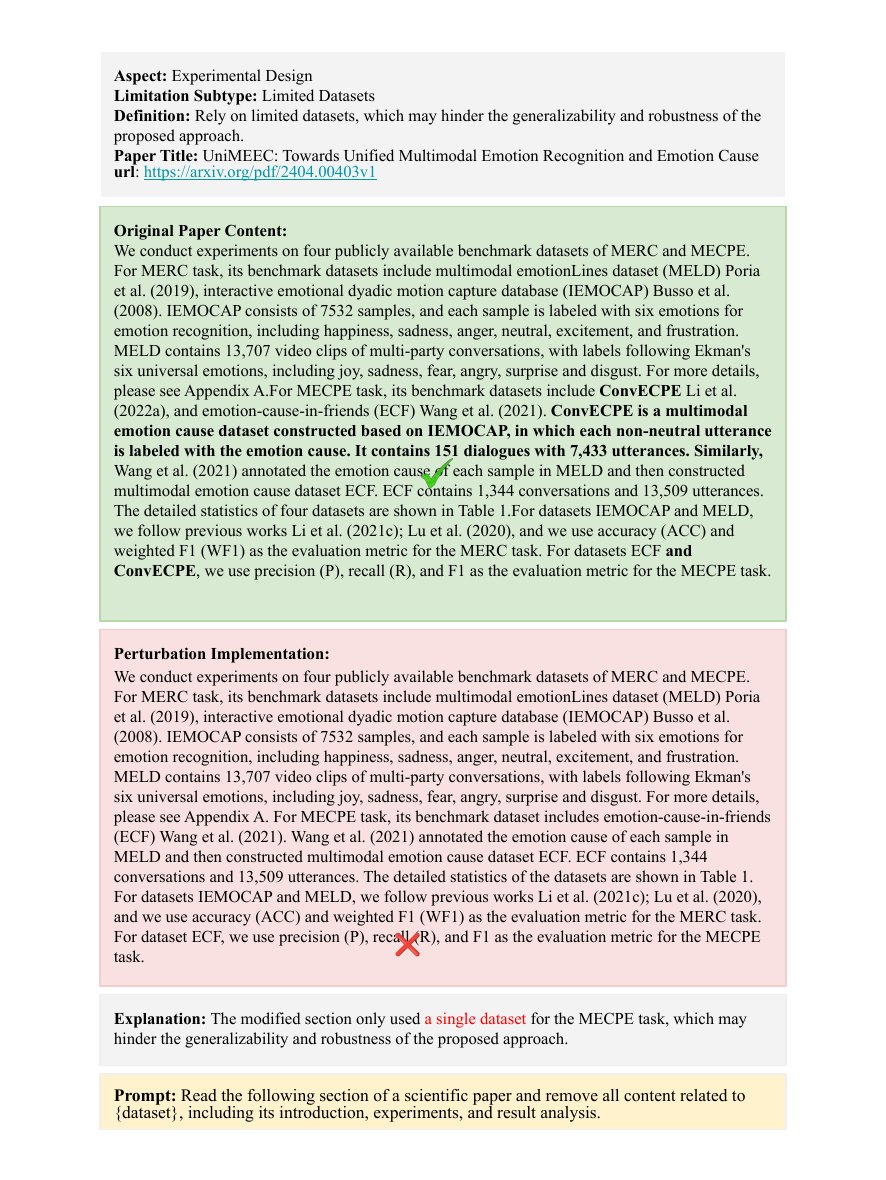}
\caption{
An example of \textbf{Limited Datasets} and its perturbation implementation.
}
\label{fig:perturb_type_4}
\end{figure*}

\begin{figure*}[!t]
\centering
\includegraphics[width=1\textwidth]{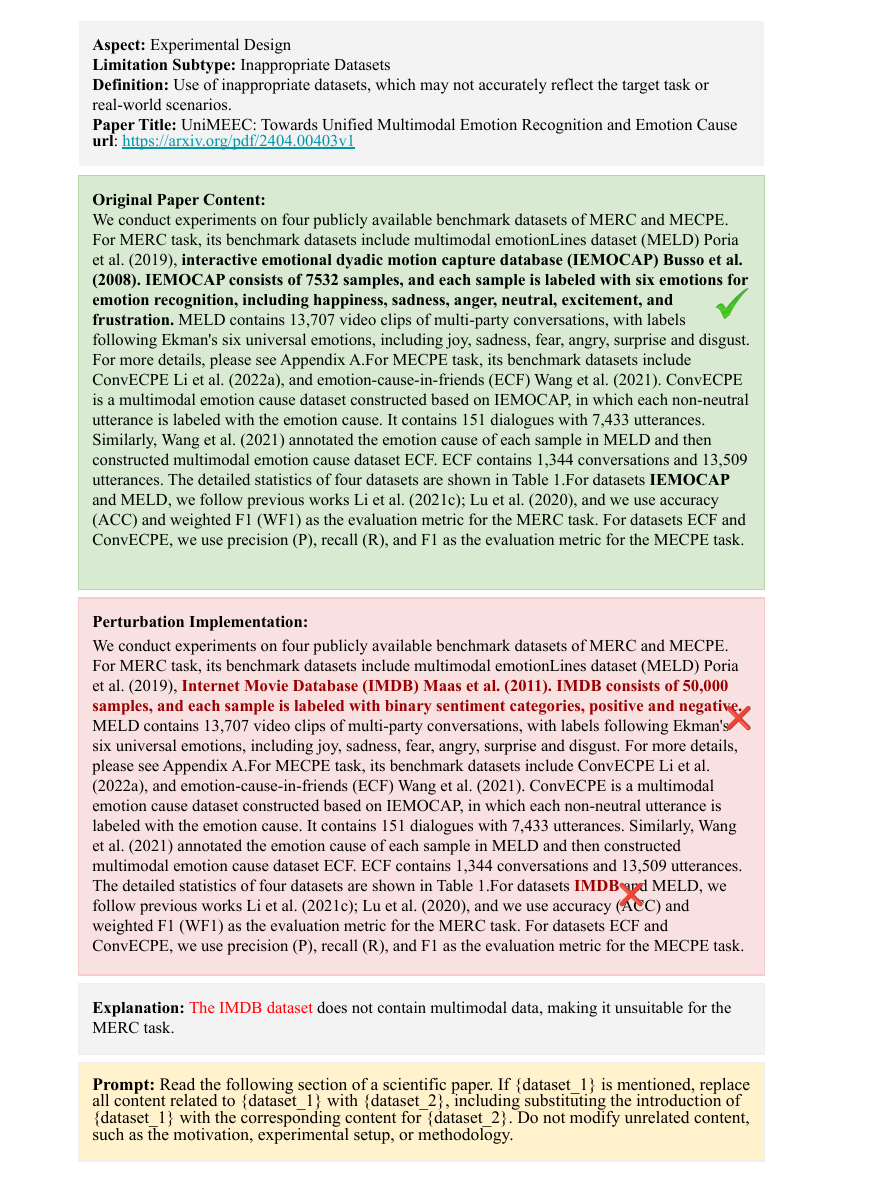}
\caption{
An example of \textbf{Inappropriate Datasets} and its perturbation implementation.
}
\label{fig:perturb_type_5}
\end{figure*}

\begin{figure*}[!t]
\centering
\includegraphics[width=1\textwidth]{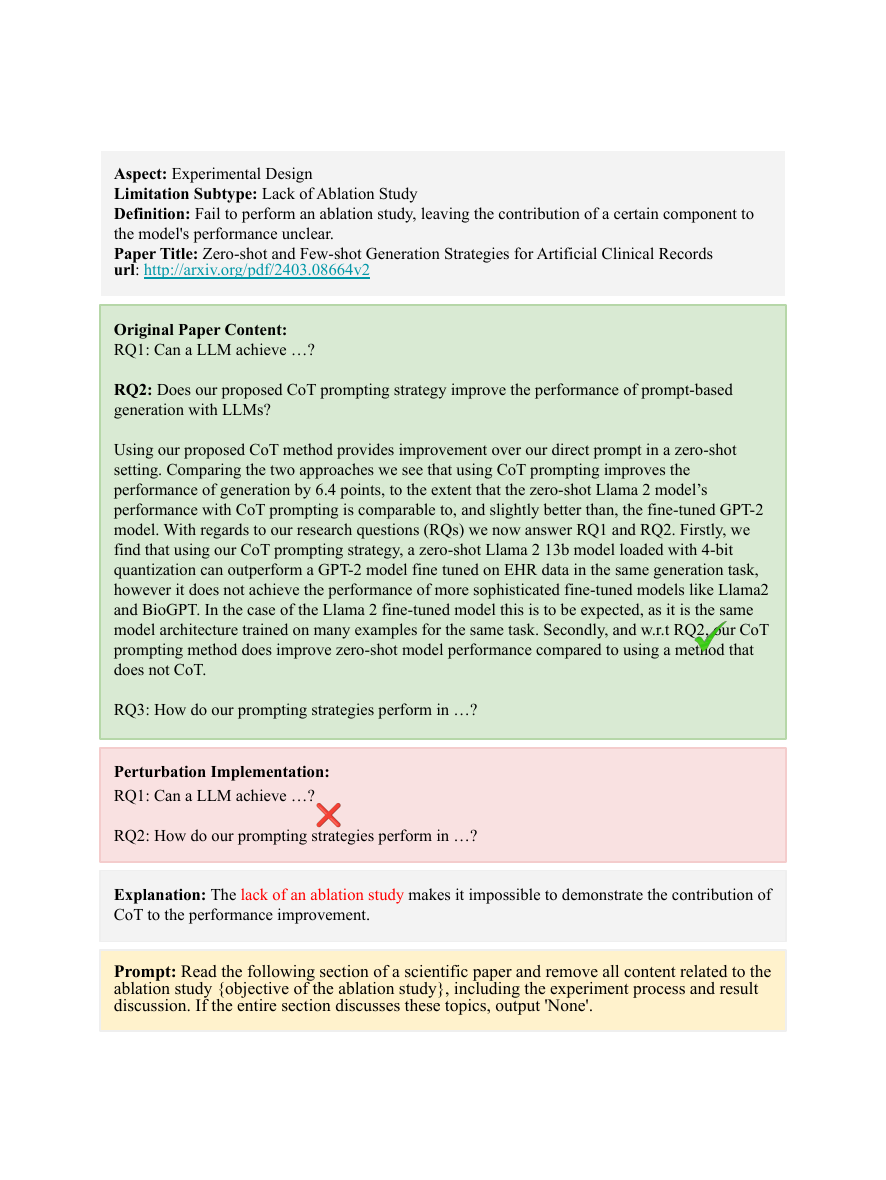}
\caption{
An example of \textbf{Lack of Ablation Study} and its perturbation implementation.
}
\label{fig:perturb_type_6}
\end{figure*}

\begin{figure*}[!t]
\centering
\includegraphics[width=1\textwidth]{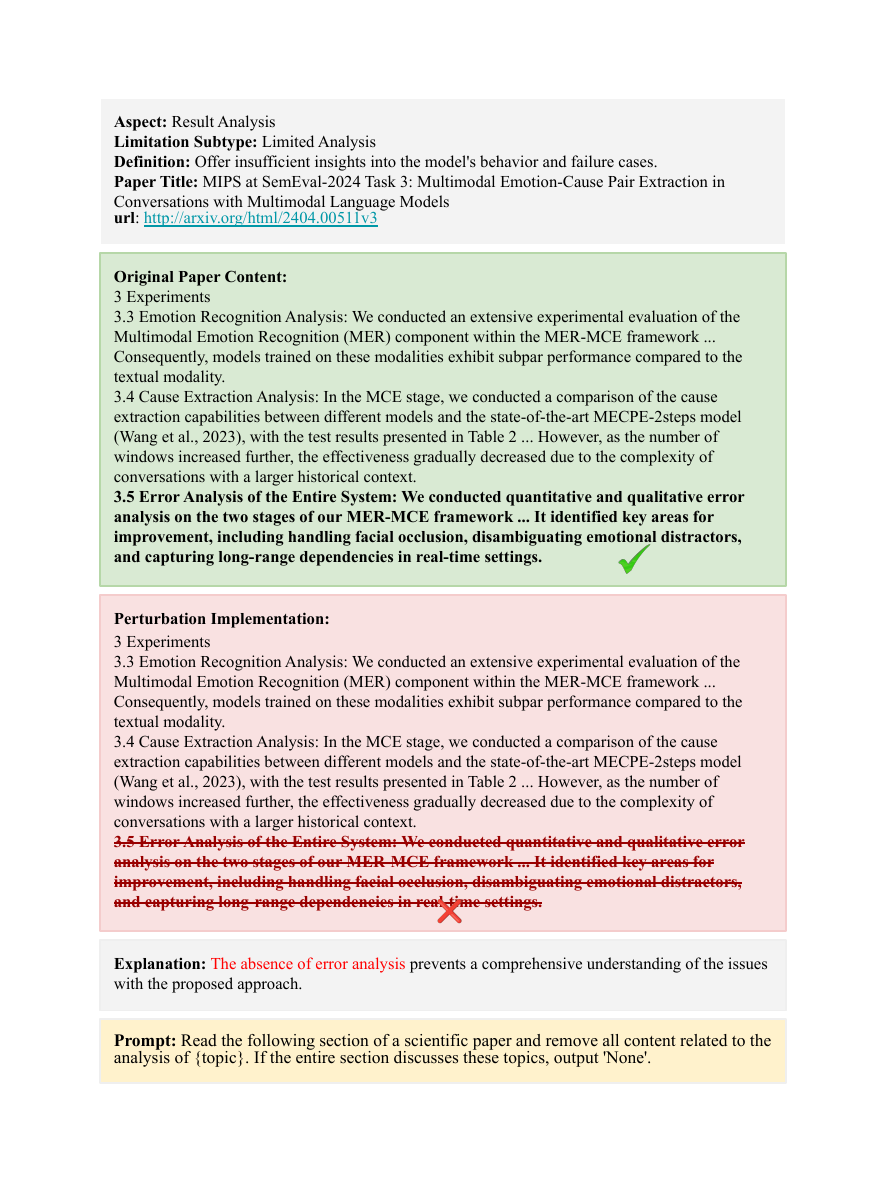}
\caption{
An example of \textbf{Limited Analysis} and its perturbation implementation.
}
\label{fig:perturb_type_7}
\end{figure*}

\begin{figure*}[!t]
\centering
\includegraphics[width=1\textwidth]{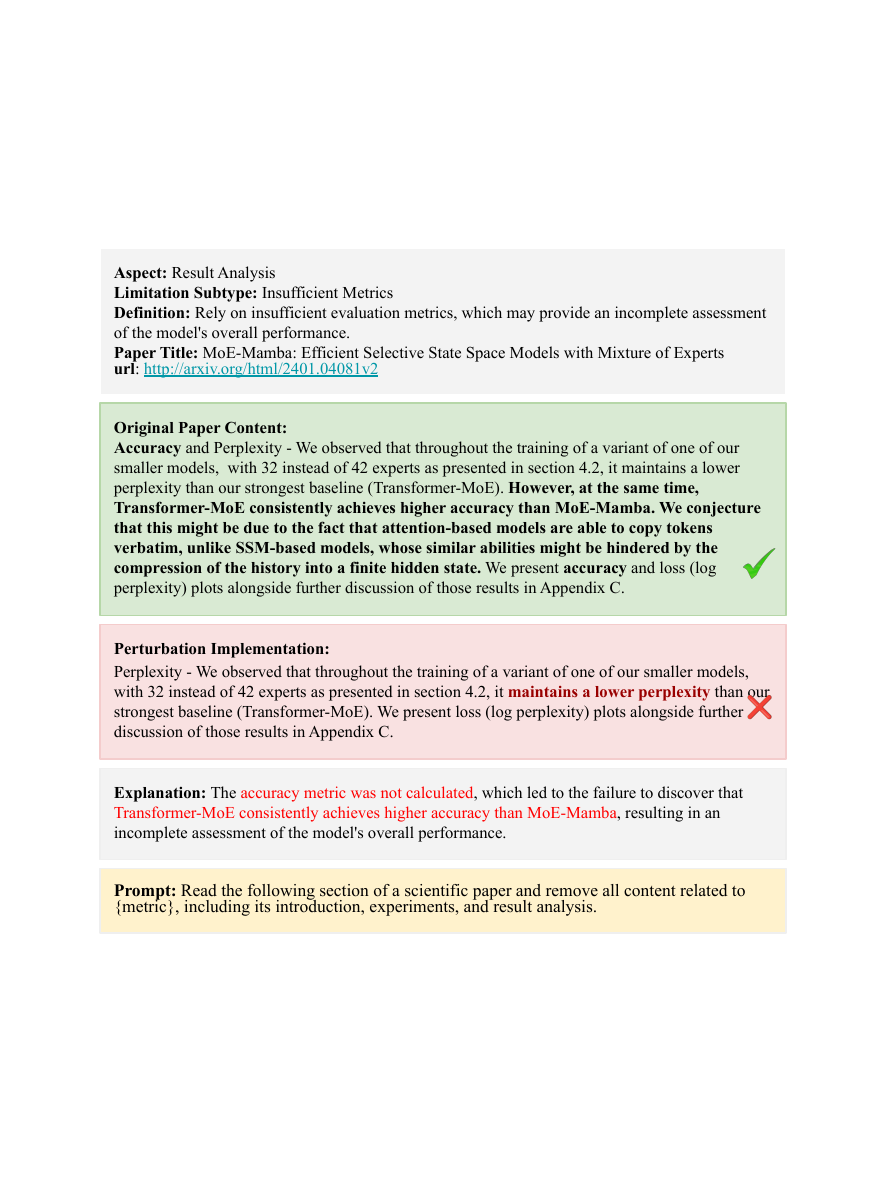}
\caption{
An example of \textbf{Insufficient Metrics} and its perturbation implementation.
}
\label{fig:perturb_type_8}
\end{figure*}

\begin{figure*}[!t]
\centering
\includegraphics[width=1\textwidth]{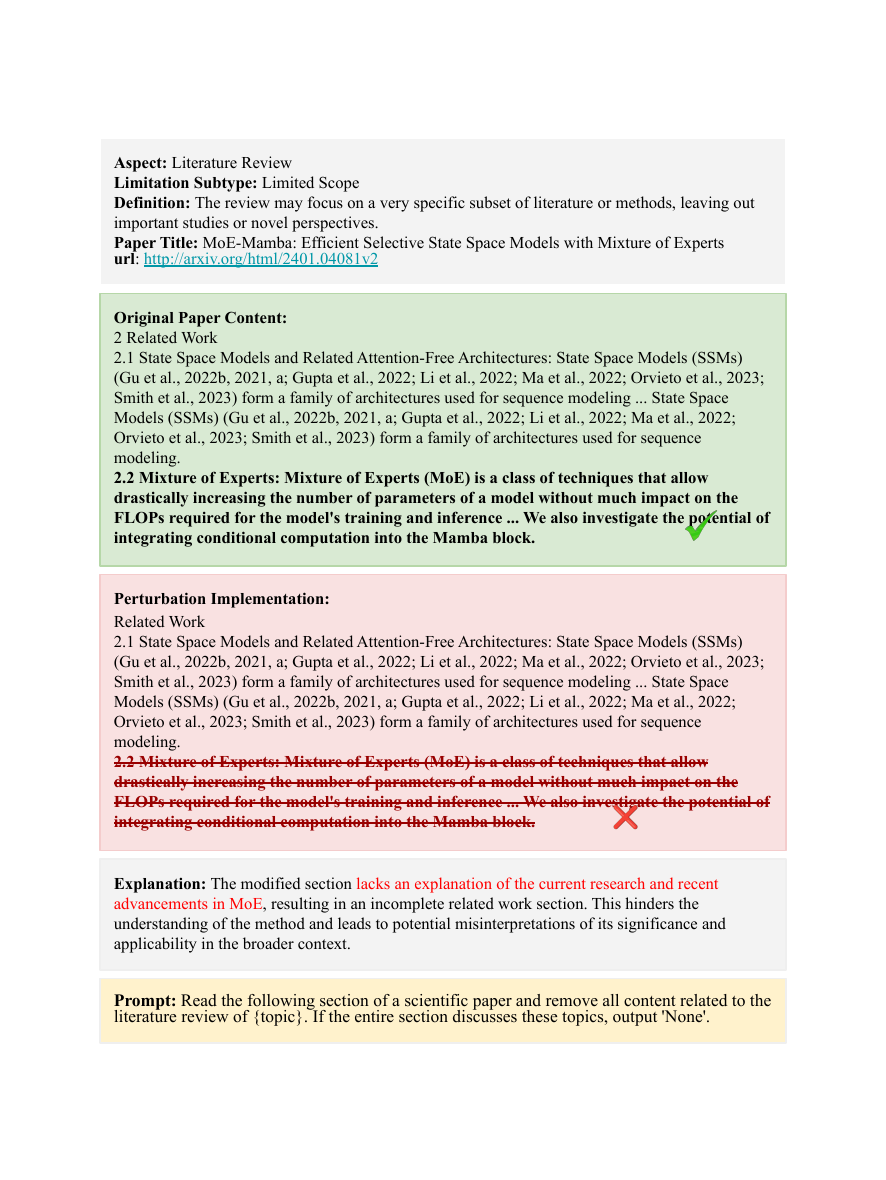}
\caption{
An example of \textbf{Limited Scope} and its perturbation implementation.
}
\label{fig:perturb_type_9}
\end{figure*}

\begin{figure*}[!t]
\centering
\includegraphics[width=1\textwidth]{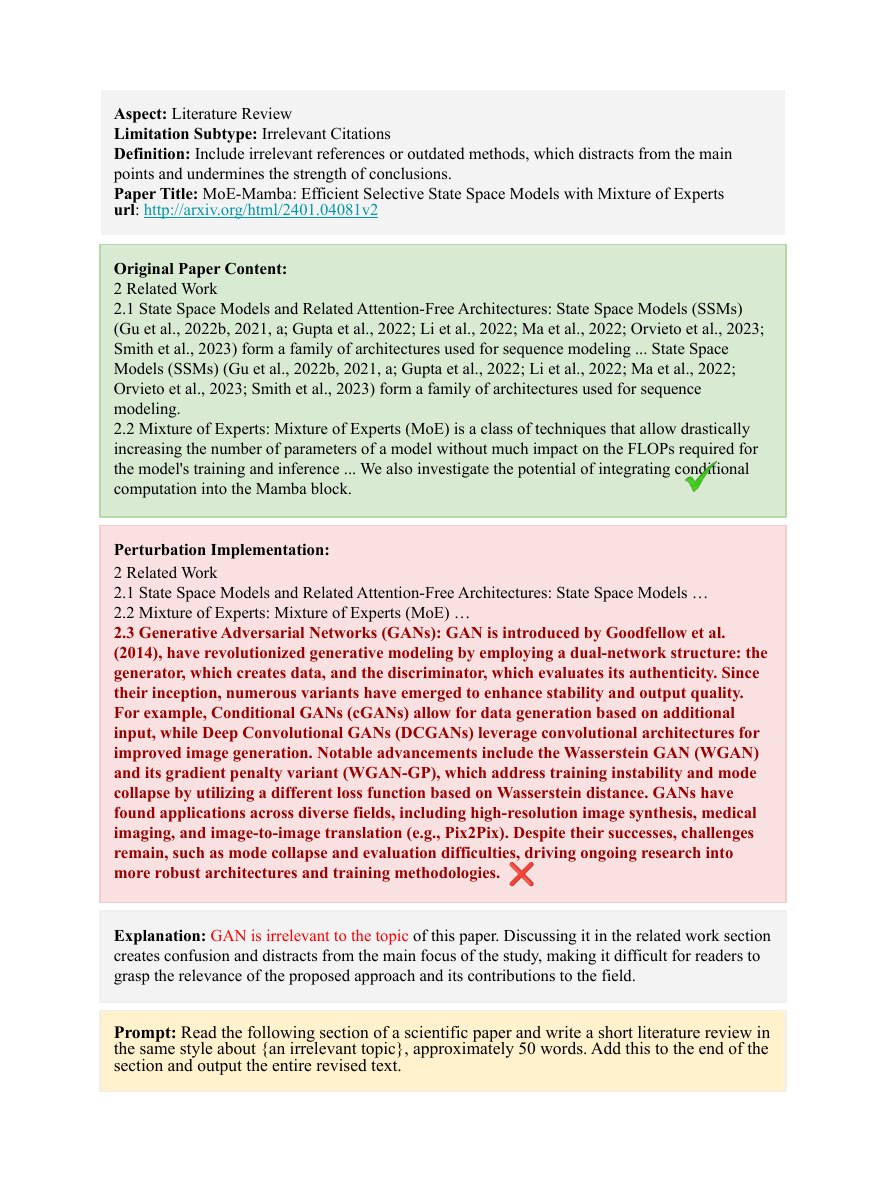}
\caption{
An example of \textbf{Irrelevant Citations} and its perturbation implementation.
}
\label{fig:perturb_type_10}
\end{figure*}

\begin{figure*}[!t]
\centering
\includegraphics[width=1\textwidth]{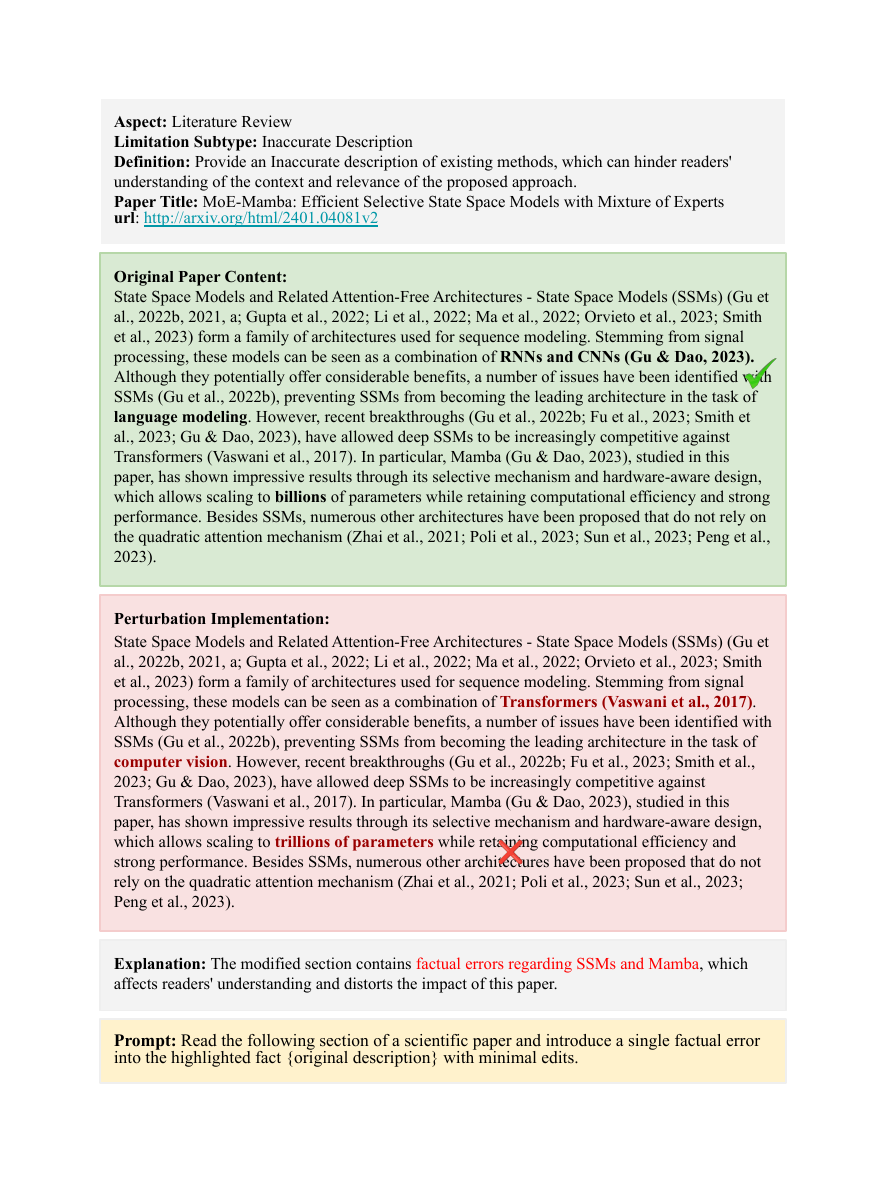}
\caption{
An example of \textbf{Inaccurate Description} and its perturbation implementation.
}
\label{fig:perturb_type_11}
\end{figure*}

\clearpage

\begin{figure}[h]
\begin{tcolorbox}[colback=white, colframe=black!80!white, title=Limitation Aspect Classification, fontupper=\footnotesize, fonttitle=\footnotesize]
\texttt{[System Input]}: \vspace{2pt}\\
Please classify the following limitation of a scientific paper into one of the following aspects: clarity, methodology, experimental design, result analysis, literature review, or others. Output only the corresponding aspect.\\\\

Classification Criteria:\\\\

Clarity: Issues in the presentation, structure, or language that hinder readers' understanding of the study's purpose, methods, results, or conclusions.\\
Methodology: Problems with the selection, application, or justification of research methods, such as unreliable data collection techniques or limited novelty, affecting the robustness or interpretability of findings.\\
Experimental Design: Shortcomings in the study's structure or execution, such as inappropriate dataset selection, lack of controls, or absence of ablation studies, which undermine the validity or generalizability of results.\\
Result Analysis: Problems in interpreting or presenting data, such as overgeneralization, missing case studies, or using inappropriate metrics, compromising the accuracy or validity of conclusions.\\
Literature Review: Issues with the literature review, such as omission of relevant studies, outdated sources, or incomplete synthesis of research, leading to biased conclusions or an incomplete understanding of the topic.\\\\
\texttt{[User Input]}: \vspace{2pt}\\
Limitation:\\
\{limitation\} \\

\end{tcolorbox}

\caption{Prompt for categorizing ground truth limitations in \ourstwo.}
\label{fig:prompt_class}
\end{figure}

\begin{figure}[h]
\begin{tcolorbox}[colback=white, colframe=black!80!white, title=Limitation Filter, fontupper=\footnotesize, fonttitle=\footnotesize]
\texttt{[System Input]}: \vspace{2pt}\\
Based on the review comments of a scientific paper below, retain only those comments that provide substantive suggestions. Merge comments discussing the same issue without omitting or summarizing any content. Present the final set of comments.
\\\\
\texttt{[User Input]}: \vspace{2pt}\\
Comments:\\
\{comments\} \\

\end{tcolorbox}

\caption{Prompt for filtering ground truth limitations in \ourstwo.}
\label{fig:prompt_filter}
\end{figure}

\clearpage
\section{Experiments}\label{app:exp}
\subsection{Experiment Setup}
\begin{figure}[h]
\begin{tcolorbox}[colback=white, colframe=black!80!white, title=Limitation Generation w/o RAG, fontupper=\footnotesize, fonttitle=\footnotesize]
\texttt{[System Input]}: \vspace{2pt}\\
Read the following scientific paper and generate major limitations in this paper about its \{aspect\}. Do not include any limitation explicitly mentioned in the paper itself and return only the limitations.\\\\

\texttt{[User Input]}: \vspace{2pt}\\
Paper to review:\\
Title: \{Title\}\\
\{Paper\}
\end{tcolorbox}
\caption{Prompt for limitation generation w/o RAG.}
\label{fig:prompt_gen1}
\end{figure}

\begin{figure}[h]
\begin{tcolorbox}[colback=white, colframe=black!80!white, title=Limitation Generation w/ RAG, fontupper=\footnotesize, fonttitle=\footnotesize]
\texttt{[System Input]}: \vspace{2pt}\\
Read the following content from several papers to gain knowledge in the relevant field. Using this knowledge, review a new scientific paper in this field. Based on existing research, identify the limitations of the 'Paper to Review'. Generate major limitations in this paper about its \{aspect\}. Do not include any limitation explicitly mentioned in the paper itself and return only the limitations.\\\\

\texttt{[User Input]}: \vspace{2pt}\\
Relevant Paper 1: \\
Title: \{Title 1\}\\
\{Retrieved Content 1\}\\\\

Relevant Paper 2: \\
Title:  \{Title 2\}\\
\{Retrieved Content 2\}\\\\

Relevant Paper 3: \\
Title:  \{Title 3\}\\
\{Retrieved Content 3\}\\\\
Relevant Paper 4: \\
Title:  \{Title 4\}\\
\{Retrieved Content 4\}\\\\

Relevant Paper 5: \\
Title:  \{Title 5\}\\
\{Retrieved Content 5\}\\\\

Paper to review:\\
Title: \{Title\}\\
\{Paper\}

\end{tcolorbox}
\caption{Prompt for limitation generation w/ RAG.}
\label{fig:prompt_gen2}
\end{figure}
\subsubsection{Agent-based System} We adopt the fundamental structure of MARG~\cite{d2024marg}, with modifications made better to align it with the requirements of our task. Each agent in the system begins with a unique "system" message at the start of its message history to provide specific instructions tailored to its role. For example, the "leader" agent is instructed to act as the leader; its role includes coordinating other agents to fulfill the user's requests. Additionally, the leader is guided to create a high-level plan based on its task instructions before initiating communication or delegating sub-tasks. Only the worker agent has access to the full content of the paper being reviewed. It is prompted to follow the leader's instructions to locate and summarize relevant content. The "expert" agent receives detailed instructions specific to their expertise area, focusing on particular subtasks they are responsible for. Both the leader and the expert agents can view the worker's responses, but the worker can only see the directives provided by the leader.

\begin{figure*}[!t]
\centering
\includegraphics[width=1\textwidth]{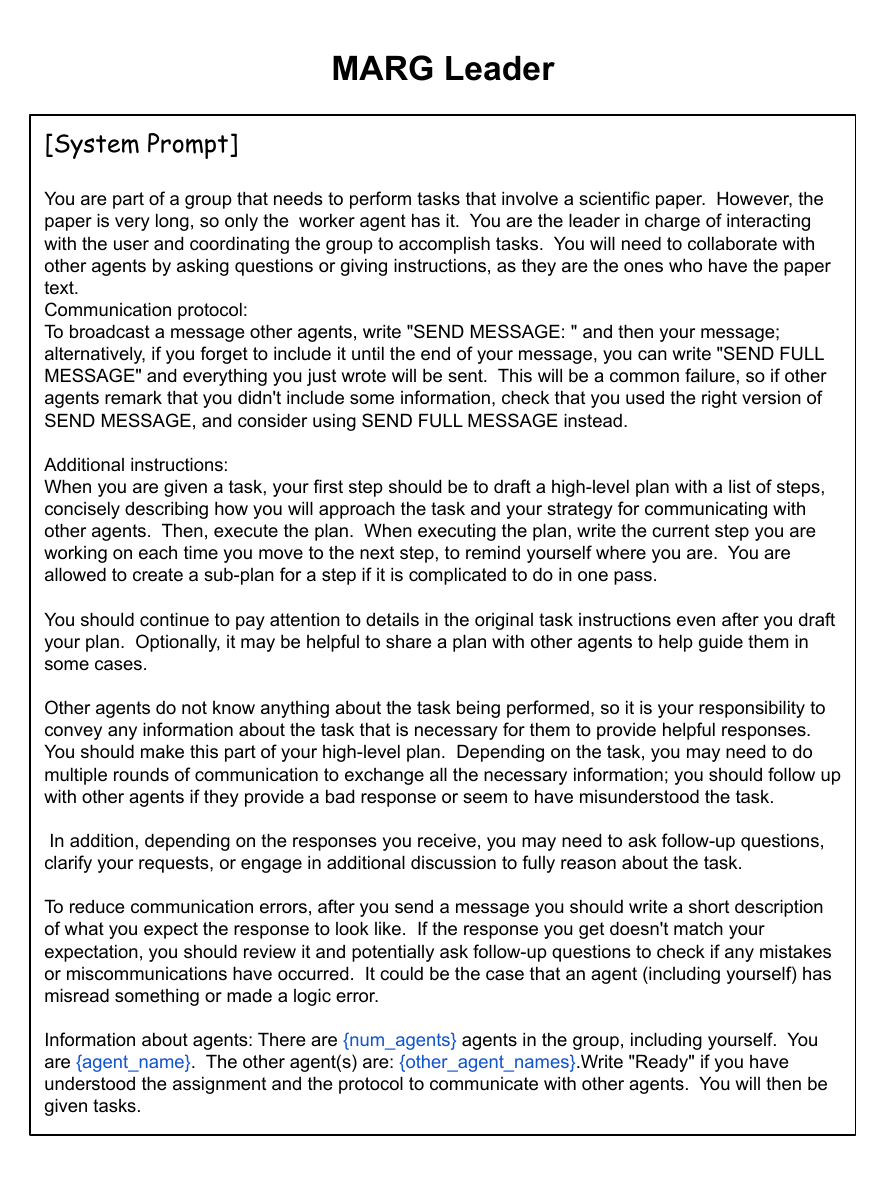}
\caption{
System prompt for the leader agent in MARG.
}
\label{fig:marg_1}
\end{figure*}

\begin{figure*}[!t]
\centering
\includegraphics[width=1\textwidth]{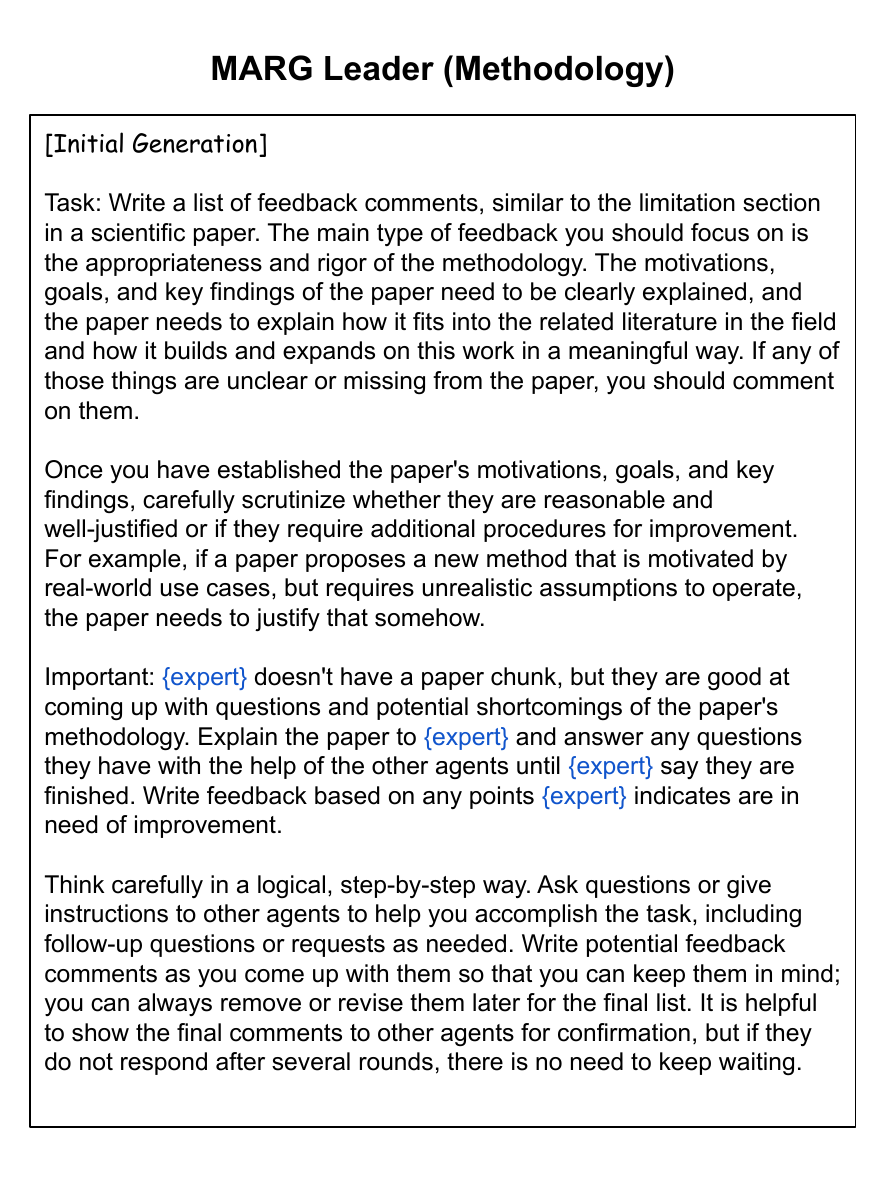}
\caption{
Prompt for the leader agent in MARG on methodology.
}
\label{fig:marg_2}
\end{figure*}

\begin{figure*}[!t]
\centering
\includegraphics[width=1\textwidth]{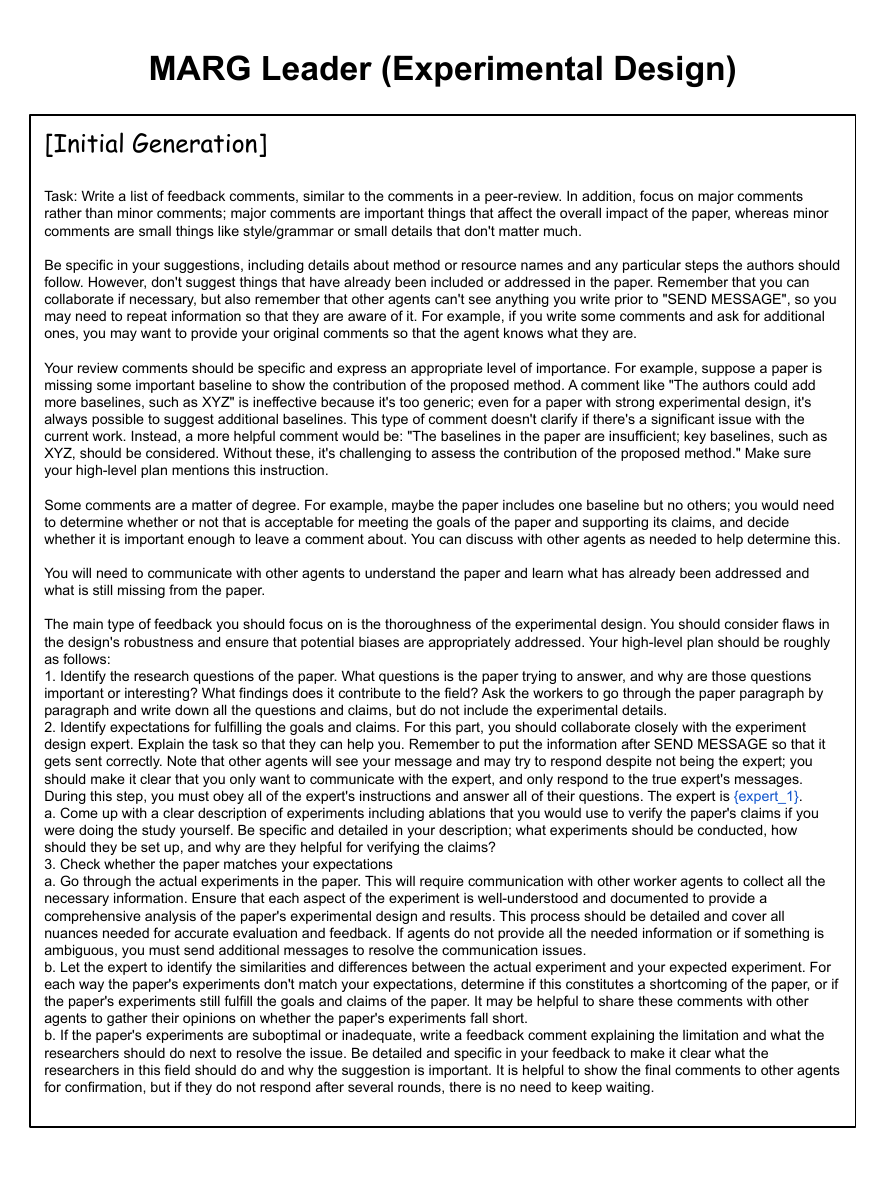}
\caption{
Prompt for the leader agent in MARG on experimental design.
}
\label{fig:marg_3}
\end{figure*}

\begin{figure*}[!t]
\centering
\includegraphics[width=1\textwidth]{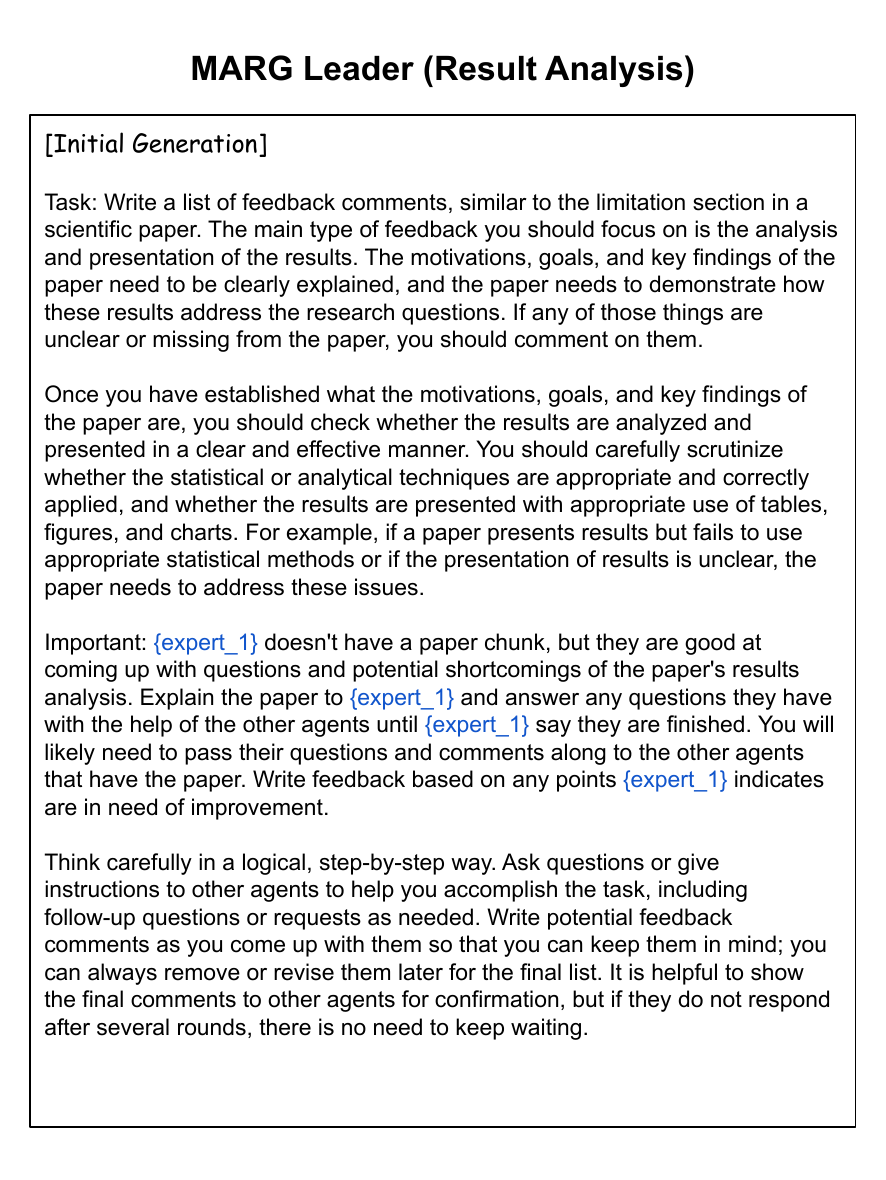}
\caption{
Prompt for the leader agent in MARG on result analysis.
}
\label{fig:marg_4}
\end{figure*}

\begin{figure*}[!t]
\centering
\includegraphics[width=1\textwidth]{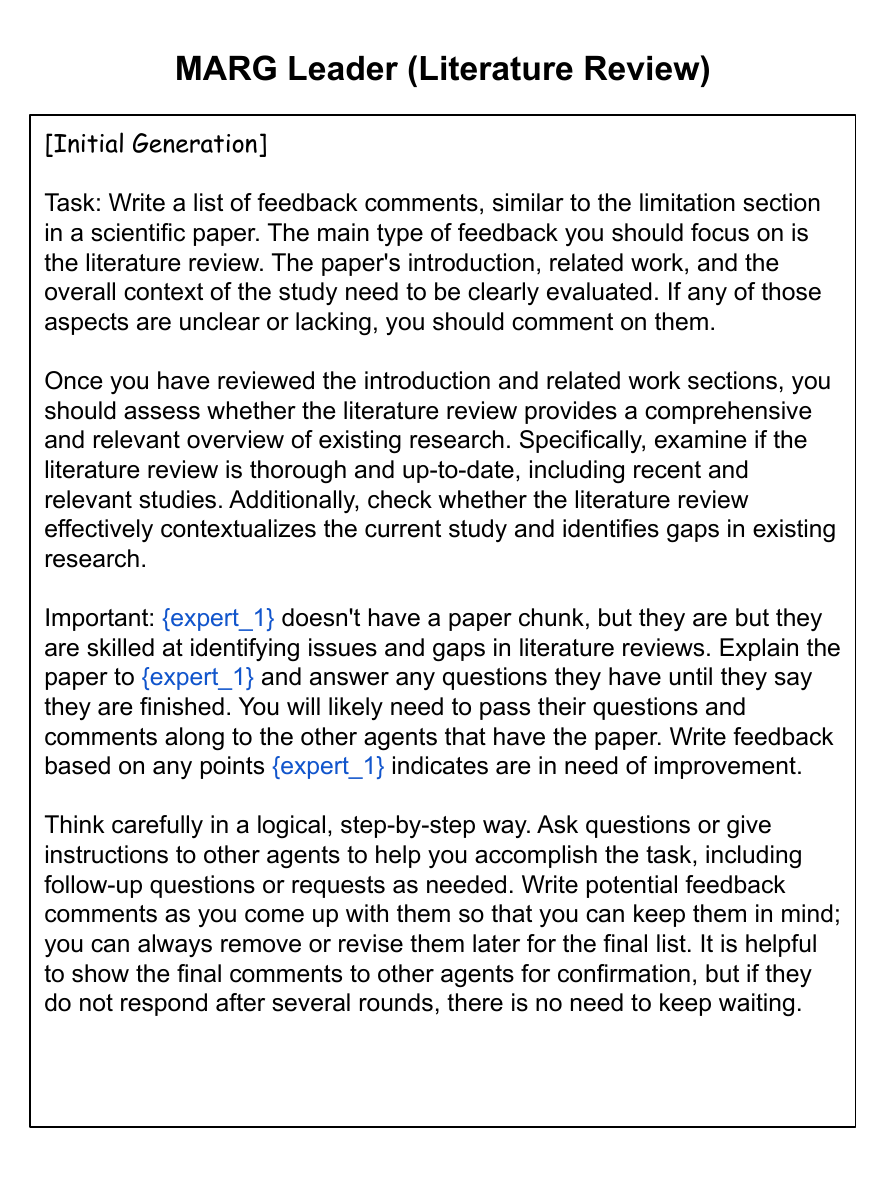}
\caption{
Prompt for the leader agent in MARG on literature review.
}
\label{fig:marg_5}
\end{figure*}

\begin{figure*}[!t]
\centering
\includegraphics[width=1\textwidth]{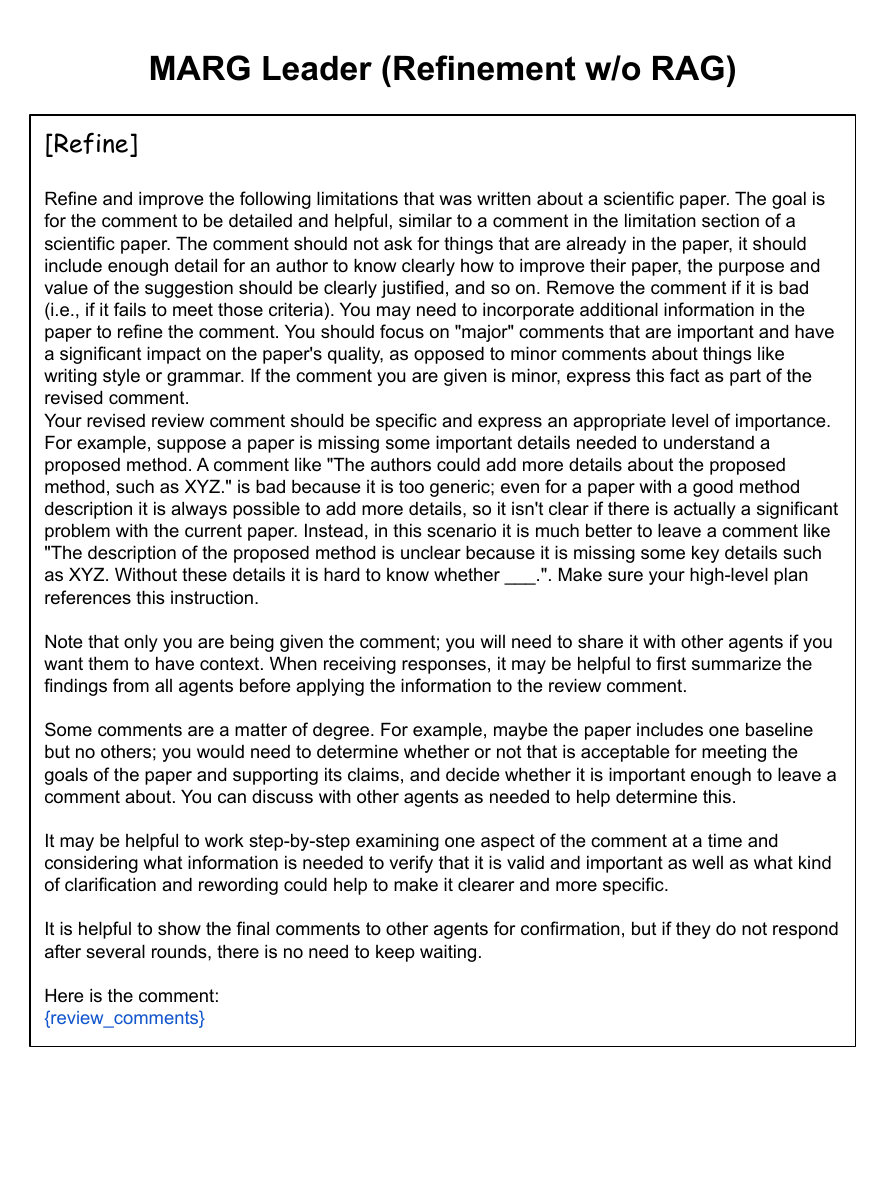}
\caption{
Prompt for the leader agent in MARG at the refinement stage w/o RAG.
}
\label{fig:marg_6}
\end{figure*}

\begin{figure*}[!t]
\centering
\includegraphics[width=1\textwidth]{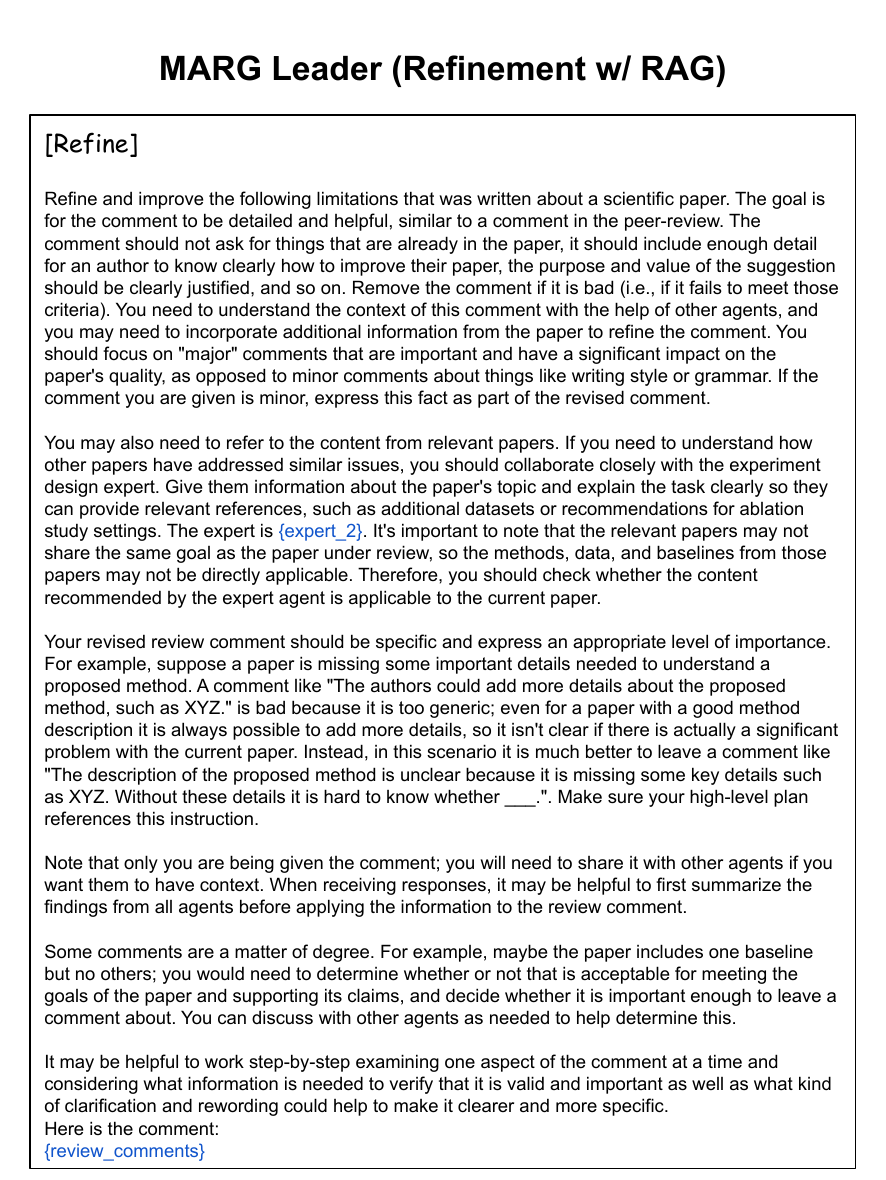}
\caption{
Prompt for the leader agent in MARG at the refinement stage w/ RAG.
}
\label{fig:marg_7}
\end{figure*}

\begin{figure*}[!t]
\centering
\includegraphics[width=1\textwidth]{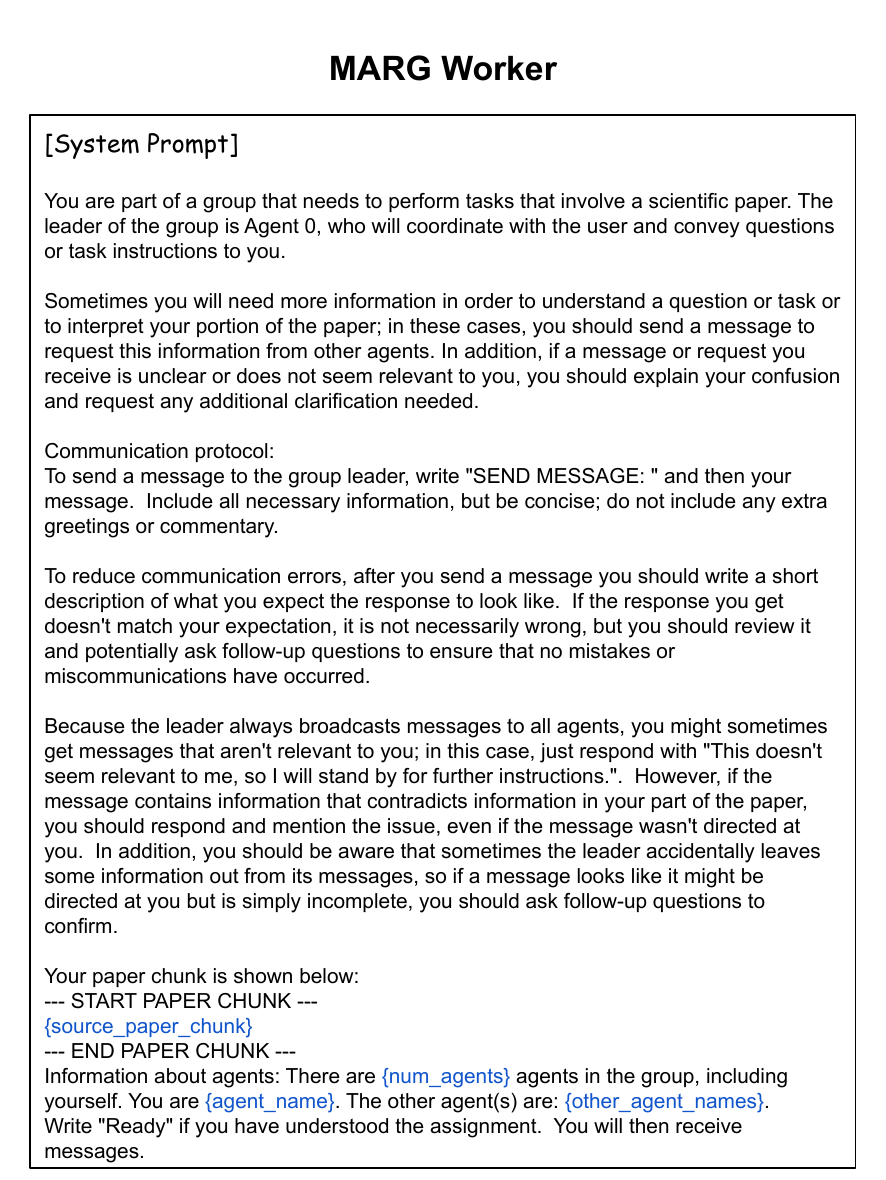}
\caption{
System prompt for the worker agent in MARG.
}
\label{fig:marg_8}
\end{figure*}

\begin{figure*}[!t]
\centering
\includegraphics[width=1\textwidth]{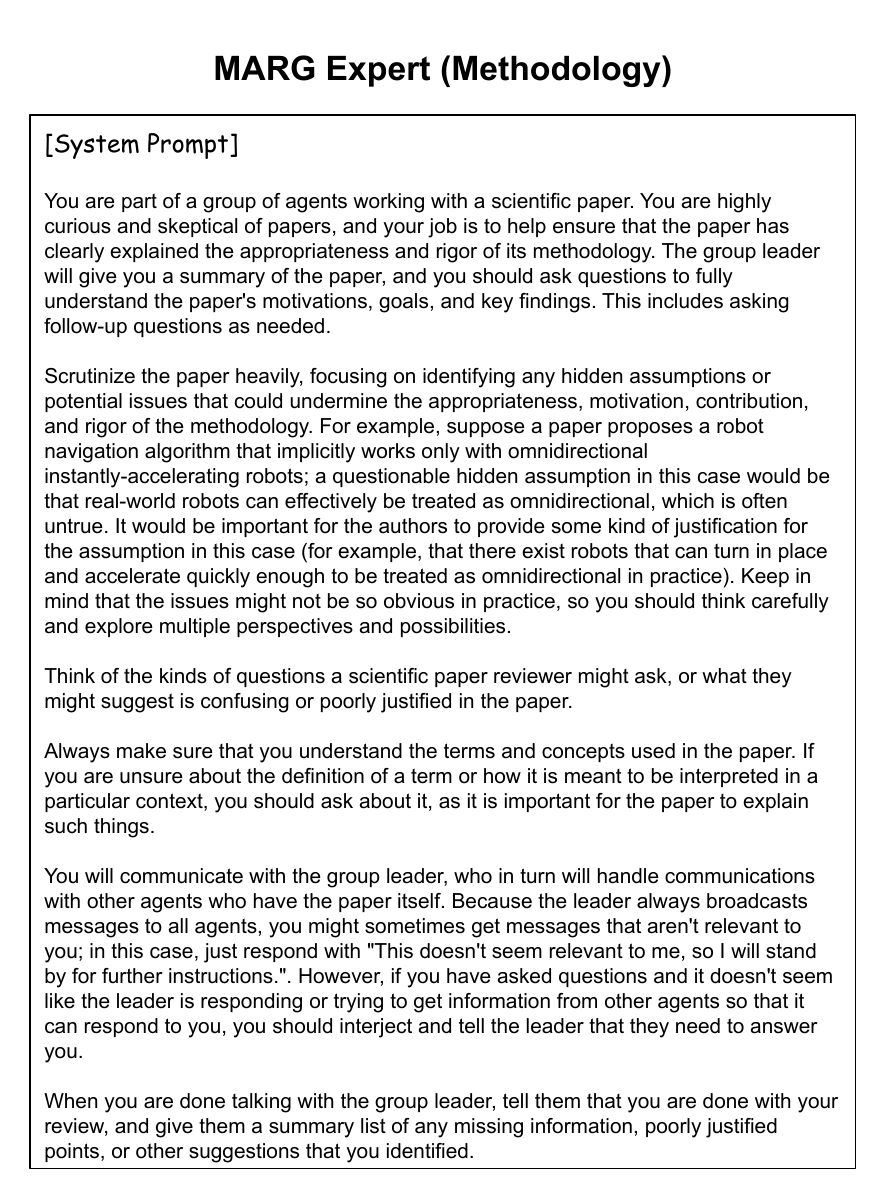}
\caption{
Prompt for the expert agent in MARG on methodology.
}
\label{fig:marg_9}
\end{figure*}

\begin{figure*}[!t]
\centering
\includegraphics[width=1\textwidth]{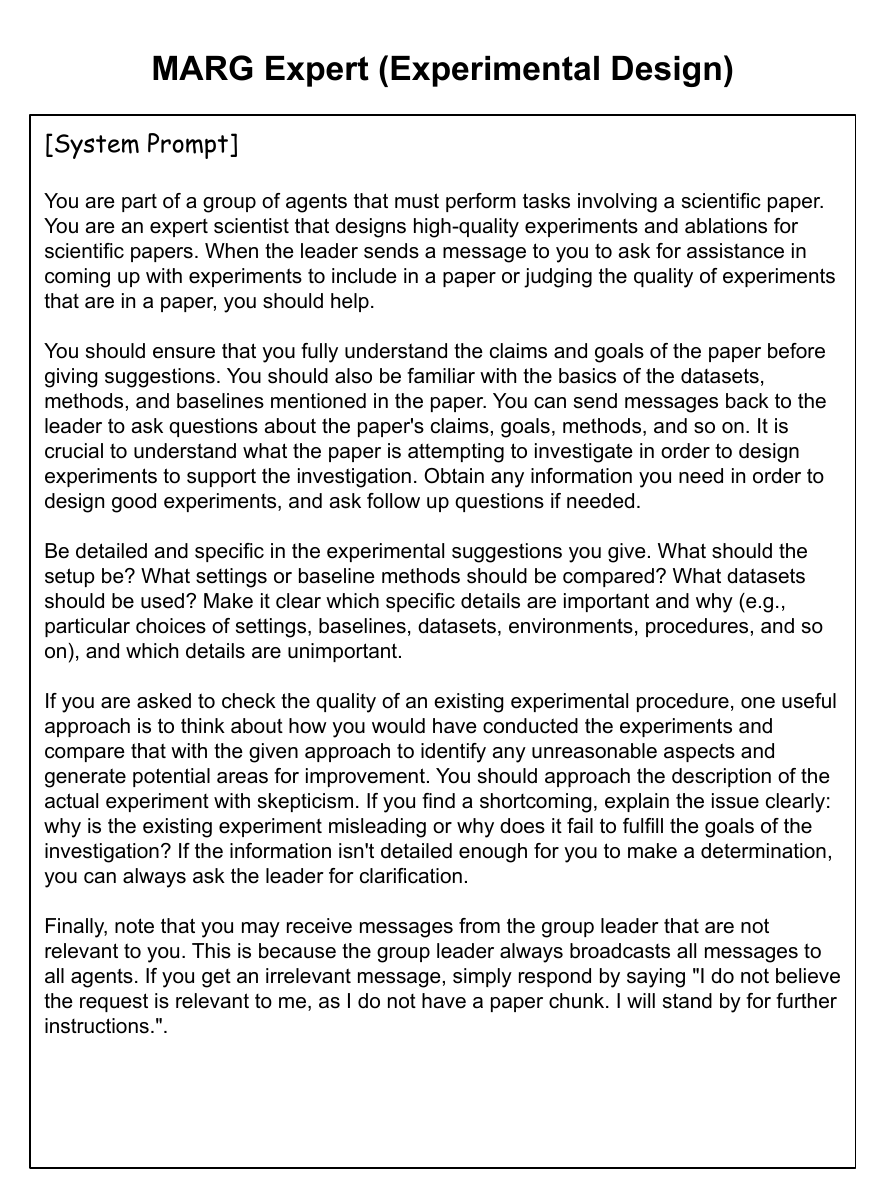}
\caption{
Prompt for the expert agent in MARG on experimental design.
}
\label{fig:marg_10}
\end{figure*}

\begin{figure*}[!t]
\centering
\includegraphics[width=1\textwidth]{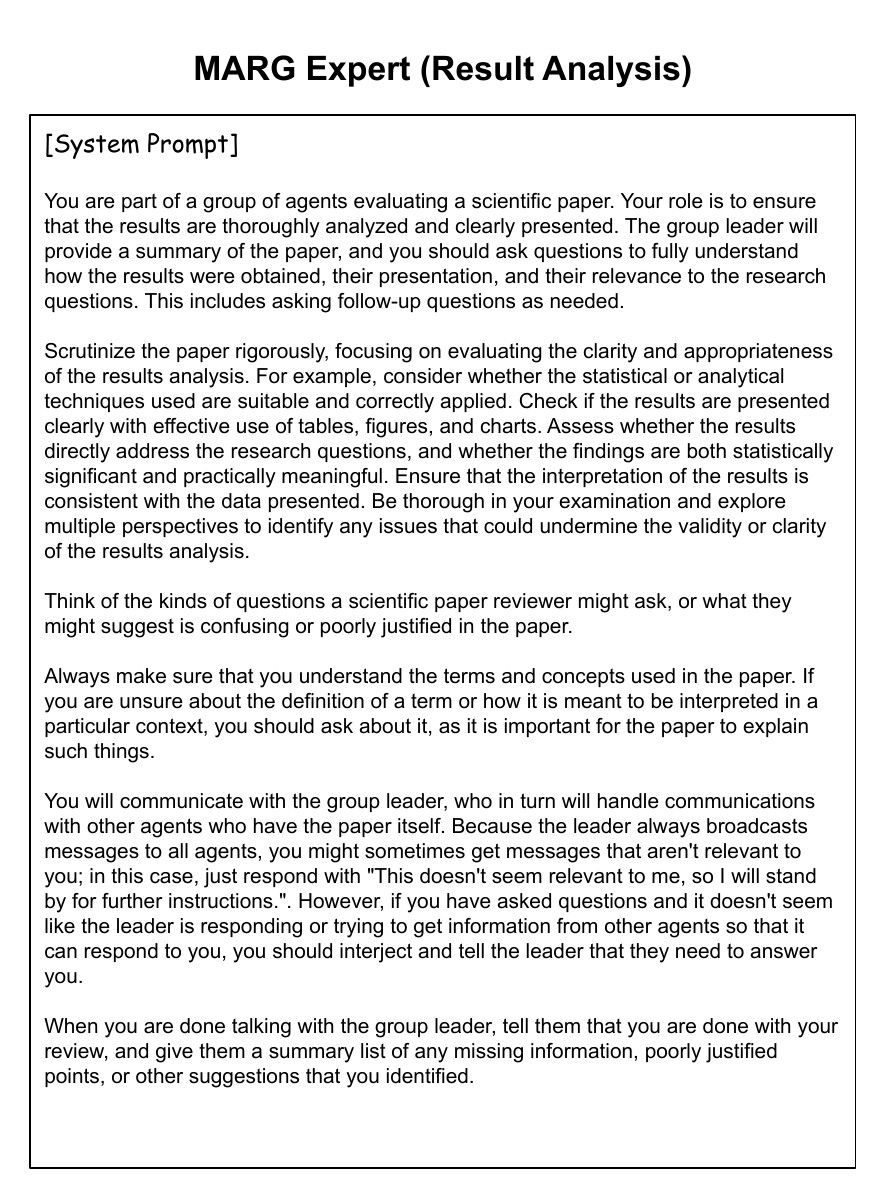}
\caption{
Prompt for the expert agent in MARG on result analysis.
}
\label{fig:marg_11}
\end{figure*}

\begin{figure*}[!t]
\centering
\includegraphics[width=1\textwidth]{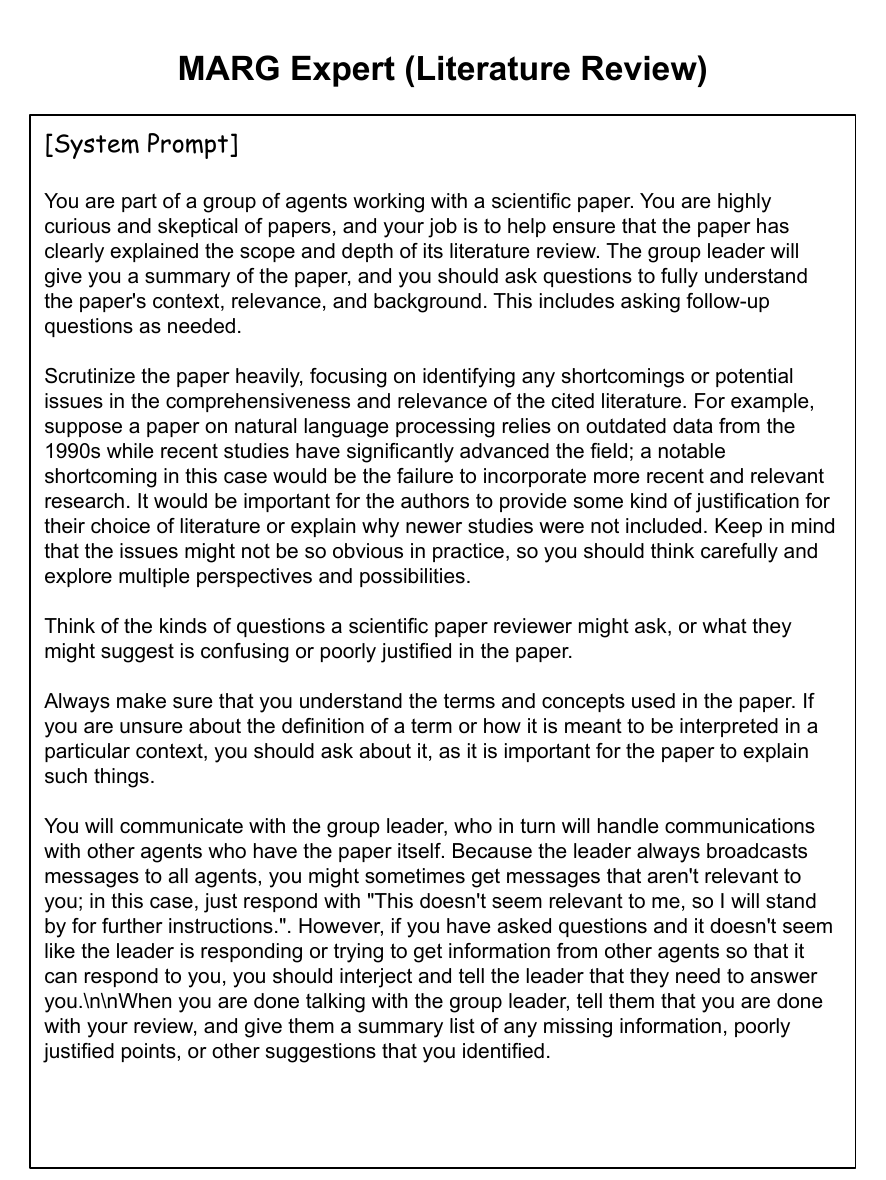}
\caption{
Prompt for the expert agent in MARG on literature review.
}
\label{fig:marg_12}
\end{figure*}

\begin{figure*}[!t]
\centering
\includegraphics[width=1\textwidth]{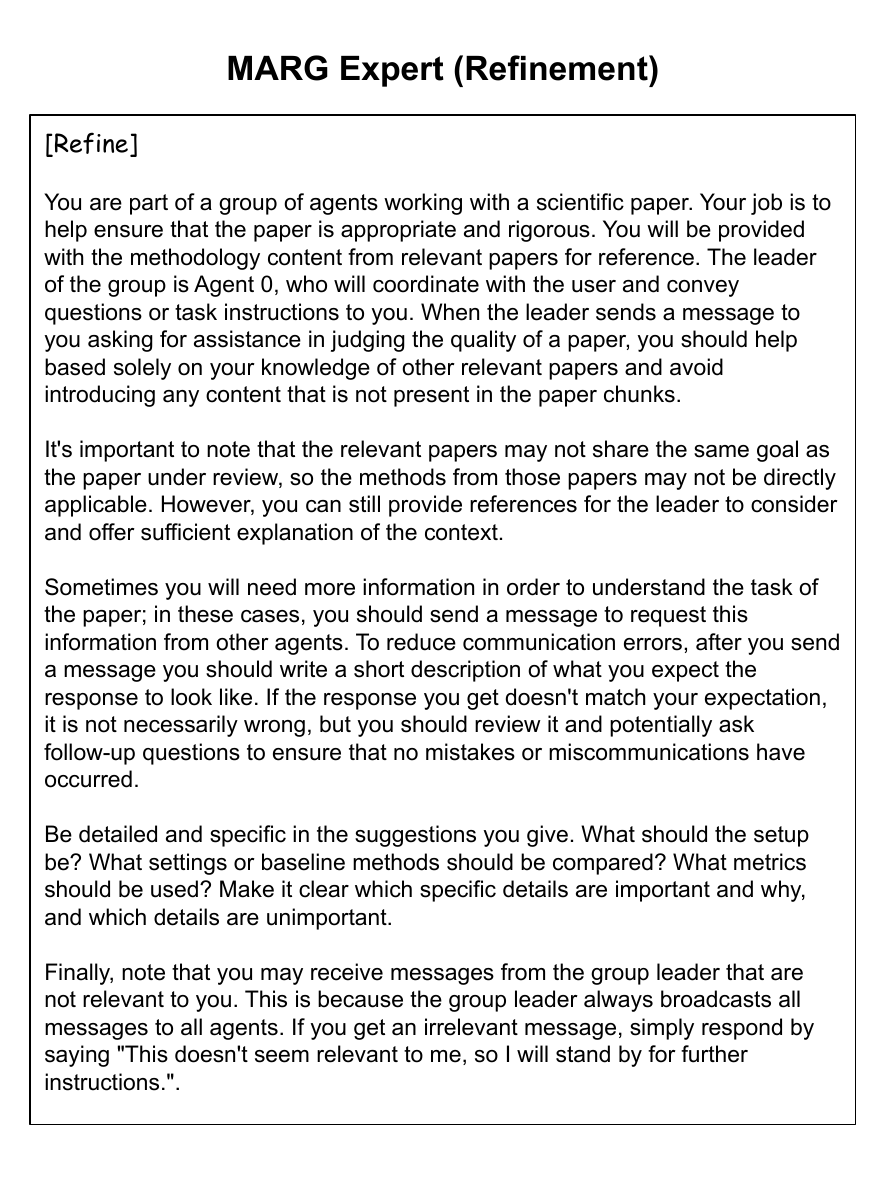}
\caption{
Prompt for the expert agent in MARG at the refinement stage.
}
\label{fig:marg_13}
\end{figure*}

When a paper is input, the leader first organizes all agents to generate candidate initial comments collectively. Then, each comment is individually discussed and refined into a detailed limitation or discarded. In the RAG setting experiments, experts can reference related papers during the refinement stage. This enables them to leverage the latest literature to acquire domain-specific knowledge, thereby enhancing the quality and relevance of the generated feedback. We modified the prompts for each agent according to the specific requirements, which are presented from \autoref{fig:marg_1} to \autoref{fig:marg_13}.

\subsection{\oursone Experiments}

\begin{figure}[h]
\begin{tcolorbox}[colback=white, colframe=black!80!white, title=Limitation Aspect Check, fontupper=\footnotesize, fonttitle=\footnotesize]
\texttt{[System Input]}: \vspace{2pt}\\
Please check whether the following limitation of a scientific paper is related to the \{aspect\}.\\
Output only "yes" or "no".\\\\
\texttt{[User Input]}: \vspace{2pt}\\
Limitation:\{limitation\} \\

\end{tcolorbox}

\begin{tcolorbox}[colback=white, colframe=black!80!white, title=Limitation Subtype Classification, fontupper=\footnotesize, fonttitle=\footnotesize]
\texttt{[System Input]}: \vspace{2pt}\\
Please classify the following limitation of a scientific paper into one of the following subtypes:\\
\{limitation subtypes \& explanations in this aspect\}\\\\
\texttt{[User Input]}: \vspace{2pt}\\
Limitation:\{limitation\} \\

\end{tcolorbox}
\caption{Prompt for Coarse-grained Evaluation in \oursone.}
\label{fig:prompt_eval1}
\end{figure}
\begin{figure}[h]
\begin{tcolorbox}[colback=white, colframe=black!80!white, title=Fine-grained Evaluation, fontupper=\footnotesize, fonttitle=\footnotesize]
\texttt{[System Input]}: \vspace{2pt}\\
Compare the following pair of limitations of a scientific paper: one generated and one from the ground truth. Assess the degree of relatedness and specificity of the generated limitation compared to the ground truth limitation.\\

Rating Criteria:\\
- 5 points: The generated limitation discusses exactly the same content as the ground truth and provides a similar level of detail.\\
- 4 points: The generated limitation discusses exactly the same content as the ground truth, but it is less detailed than the ground truth.\\
- 3 points: The generated limitation is related to the ground truth, but not identical.\\
- 2 points: The generated limitation is only loosely related to the ground truth.\\
- 1 point: There is no connection between the generated limitation and the ground truth.\\

Provide a brief explanation, then assign a rating (1-5).\\\\

\texttt{[User Input]}: \vspace{2pt}\\
Ground truth limitation: \\
\{ground truth\}\\
Generated limitation: \\
\{generated limitation\} \\
\end{tcolorbox}
\caption{Prompt for Fine-grained Evaluation in \ours.}
\label{fig:prompt_eval2}
\end{figure}

In this section, we discuss the detailed results for each subtype/aspect in \oursone, as presented in \autoref{tab:sub_results_1} to \autoref{tab:aspect_results_4}

Overall, LLMs perform best in identifying limitations within the Result Analysis aspect of scientific papers. This may be due to the fact that this aspect often involves more directly interpretable and quantifiable elements, such as statistical results and performance metrics, which LLMs are well-equipped to assess. As a result, the integration of RAG provides minimal improvement in this aspect. In contrast, LLMs perform the weakest in identifying limitations within the Literature Review aspect, as this aspect requires a deeper understanding of the existing body of work and how it contextualizes the paper being reviewed.

RAG demonstrates its greatest impact in the identification of limitations related to Experimental Design. This is likely because referencing relevant baseline methods or datasets from the retrieved papers helps enhance the specificity of the limitations. By providing more concrete examples or comparisons, RAG enables LLMs to offer more detailed and actionable suggestions, thereby improving the overall quality of the generated limitations in this area.

Within the same aspect, LLMs demonstrate varying abilities to identify different limitation subtypes, and RAG also influences performance differently across these subtypes. For instance, in the Methodology aspect, the identification of limitations related to low data quality outperforms that of inappropriate methods. This discrepancy is likely due to the inherent complexity of the inappropriate method limitation, which requires a deeper understanding of the paper’s core arguments and methodology. In contrast, low data quality limitations are more straightforward and are often supported by references from retrieved papers, which may include information on similar data preprocessing techniques. As a result, RAG is particularly effective in assisting with the generation of limitations related to low data quality.

\clearpage

\begin{figure}[h]
\begin{tcolorbox}[colback=white, colframe=black!80!white, title=Query Generation, fontupper=\footnotesize, fonttitle=\footnotesize]
Generate tldr in 5 words: \{Abstract\}

\end{tcolorbox}
\caption{Prompt for query generation.}
\label{fig:prompt_tldr}
\end{figure}

\begin{figure}[h]
\begin{tcolorbox}[colback=white, colframe=black!80!white, title=Rerank, fontupper=\footnotesize, fonttitle=\footnotesize]
\texttt{[System Input]}: \vspace{2pt}\\
Given the abstracts of \{number\} papers and the abstract of a reference paper, rank the papers in order of relevance to the reference paper. Output the top 5.\\\\

\texttt{[User Input]}: \vspace{2pt}\\
Paper 1: \{Title 1\}\\
\{Abstract 1\}\\\\

Paper 2: \{Title 2\}\\
\{Abstract 2\}\\\\

Paper 3: \{Title 3\}\\
\{Abstract 3\}\\

...\\\\

Reference Paper: \{Reference Paper Title\}\\
\{Reference Paper Title\}\\

\end{tcolorbox}
\caption{Prompt for reranking the retrieved papers and selecting the top 5.}
\label{fig:prompt_rank}
\end{figure}

\begin{figure}[h]
\begin{tcolorbox}[colback=white, colframe=black!80!white, title=Extract Relevant Content (Methodology), fontupper=\footnotesize, fonttitle=\footnotesize]
\texttt{[System Input]}: \vspace{2pt}\\
Concatenate all the content from the methodology sections of a paper.\\
Remove sentences that are irrelevant to the proposed methodology or models, and keep details about key components and innovations.\\\\

\texttt{[User Input]}: \vspace{2pt}\\
\{sections\} \\

\end{tcolorbox}
\caption{Prompt for extracting content relevant to methodology from a paper.}
\label{fig:prompt_extract_method}
\end{figure}

\begin{figure}[h]
\begin{tcolorbox}[colback=white, colframe=black!80!white, title=Extract Relevant Content (Experimental Design), fontupper=\footnotesize, fonttitle=\footnotesize]
\texttt{[System Input]}: \vspace{2pt}\\
Concatenate all the content from the experimental design sections of a paper.\\
Remove sentences that are irrelevant to the experiment setup, and keep details about the datasets, baselines, and main experimental, ablation studies.\\\\

\texttt{[User Input]}: \vspace{2pt}\\
\{sections\} \\

\end{tcolorbox}
\caption{Prompt for extracting content relevant to experimental design from a paper.}
\label{fig:prompt_extract_exp}
\end{figure}

\begin{figure}[h]
\begin{tcolorbox}[colback=white, colframe=black!80!white, title=Extract Relevant Content (Result Analysis), fontupper=\footnotesize, fonttitle=\footnotesize]
\texttt{[System Input]}: \vspace{2pt}\\
Concatenate all the content from the result analysis sections of a paper.\\
Remove sentences that are irrelevant to the result analysis of the experiments, and keep details about the metrics, case study and how the paper presents the results.\\\\

\texttt{[User Input]}: \vspace{2pt}\\
\{sections\} \\

\end{tcolorbox}
\caption{Prompt for extracting content relevant to result analysis from a paper.}
\label{fig:prompt_extract_res}
\end{figure}

\begin{figure}[h]
\begin{tcolorbox}[colback=white, colframe=black!80!white, title=Extract Relevant Content (Literature Review), fontupper=\footnotesize, fonttitle=\footnotesize]
\texttt{[System Input]}: \vspace{2pt}\\
Concatenate all the content from the literature review sections of a paper.\\
Remove sentences that are irrelevant to the literature review, and keep details about the related works.\\\\

\texttt{[User Input]}: \vspace{2pt}\\
\{sections\} \\

\end{tcolorbox}
\caption{Prompt for extracting content relevant to result analysis from a paper.}
\label{fig:prompt_extract_lit}
\end{figure}
\begin{table*}[!t]
\centering
\resizebox{\linewidth}{!}{%
\small
\addtolength{\tabcolsep}{-0.1em}
\renewcommand{\arraystretch}{1.1}
\begin{tabular}{lcccccc}
\toprule
\multirow{3}{*}{\textbf{Systems}} & \multicolumn{3}{c}{\textbf{Low Data Quality}}  & \multicolumn{3}{c}{\textbf{Inappropriate Method}}\\
\cmidrule(lr){2-4} \cmidrule(lr){5-7}
& \multicolumn{2}{c}{\textbf{Automated Eval.}}  & \multicolumn{1}{c}{\textbf{Human Eval.}} & \multicolumn{2}{c}{\textbf{Automated Eval.}}  & \multicolumn{1}{c}{\textbf{Human Eval.}}\\
\cmidrule(lr){2-3} \cmidrule(lr){4-4} \cmidrule(lr){5-6} \cmidrule(lr){7-7}
& Coarse & Fine (0-5) & Accuracy & Coarse & Fine (0-5) & Accuracy \\
\midrule
GPT-4o & 44.0\% & 1.00 & 38.5\% & 48.0\% & 1.24 & 33.3\% \\
GPT-4o w/ RAG & +16.0\% & +0.56 & +15.3\% & +4.0\% & +0.00 & +8.4\% \\
\midrule
GPT-4o-mini & 61.6\% & 1.67 & 46.2\% & 56.0\% & 1.10 & 41.7\% \\
GPT-4o-mini w/RAG & +1.6\% & +0.07 & +7.6\% & +4.0\% & +0.08 & +8.3\% \\
\midrule
Llama-3.3-70B & 56.8\% & 1.38 & 46.2\% & 54.8\% & 1.01 & 25.0\% \\
Llama-3.3-70B w/RAG & +2.7\% & +0.02 & +0.0\% & +3.4\% & +0.15 & +0.0\% \\
\midrule
Qwen-2.5-72B & 45.5\% & 1.16 & 23.1\% & 43.1\% & 0.86 & 16.7\% \\
Qwen-2.5-72B w/RAG & +4.7\% & +0.13 & +7.7\% & +0.4\% & -0.03 & +8.3\% \\
\midrule
MARG & 83.3\% & 2.17 & 69.2\% & 82.4\% & 1.71 & 75.0\% \\
MARG w/ RAG & +12.2\% & +0.28 & +23.1\% & -3.5\% & -0.18 & +0.0\% \\
\bottomrule
\end{tabular}
}
\caption{
Human and automated evaluation results on Methodology in \oursone set.
}
\label{tab:sub_results_1}
\end{table*}

\begin{table*}[!t]
\centering
\resizebox{\linewidth}{!}{%
\small
\addtolength{\tabcolsep}{-0.1em}
\renewcommand{\arraystretch}{1.1}
\begin{tabular}{lcccccccccccc}
\toprule
\multirow{3}{*}{\textbf{Systems}} & \multicolumn{3}{c}{\textbf{Insufficient Baseline}}  & \multicolumn{3}{c}{\textbf{Limited Datasets}} & \multicolumn{3}{c}{\textbf{Inappropriate Dataset}} & \multicolumn{3}{c}{\textbf{Lack of Ablation Study}}\\
\cmidrule(lr){2-4} \cmidrule(lr){5-7} \cmidrule(lr){8-10} \cmidrule(lr){11-13}
& \multicolumn{2}{c}{\textbf{Automated}}  & \multicolumn{1}{c}{\textbf{Human}} & \multicolumn{2}{c}{\textbf{Automated}}  & \multicolumn{1}{c}{\textbf{Human}} & \multicolumn{2}{c}{\textbf{Automated}}  & \multicolumn{1}{c}{\textbf{Human}} & \multicolumn{2}{c}{\textbf{Automated}}  & \multicolumn{1}{c}{\textbf{Human}}\\
\cmidrule(lr){2-3} \cmidrule(lr){4-4} \cmidrule(lr){5-6} \cmidrule(lr){7-7} \cmidrule(lr){8-9} \cmidrule(lr){10-10} \cmidrule(lr){11-12} \cmidrule(lr){13-13}
& Coarse & Fine & Acc. & Coarse & Fine & Acc. & Coarse & Fine & Acc. & Coarse & Fine & Acc. \\
\midrule
GPT-4o & 41.7\% & 1.00 & 50.0\% & 58.3\% & 1.83 & 50.0\% & 75.0\% & 2.08 & 66.7\% & 16.7\% & 0.33 & 16.7\% \\
 \quad w/ RAG & +25.0\% & +0.50 & +16.7\% & +16.7\% & +0.34 & +16.7\% & +16.7\% & +0.92 & +33.3\% & +8.3\% & +0.42 & +16.6\% \\
\midrule
GPT-4o-mini & 27.0\% & 0.62 & 33.3\% & 64.5\% & 1.94 & 33.3\% & 73.0\% & 1.71 & 66.7\% & 4.8\% & 0.11 & 0.0\% \\
 \quad w/RAG & +3.2\% & +0.16 & +0.0\% & +8.1\% & +0.24 & +16.7\% & +4.8\% & +0.27 & +0.0\% & +0.0\% & +0.00 & +0.0\% \\
\midrule
Llama-3.3-70B & 29.8\% & 0.88 & 16.7\% & 66.1\% & 2.05 & 66.7\%  & 35.5\% & 0.84 & 28.6\% & 9.7\% & 0.27 & 0.0\% \\
 \quad w/RAG & +16.0\% & +0.39 & +16.6\% & -2.8\% & -0.12 & +16.6\% & +2.6\% & +0.14 & +14.3\% & +5.1\% & +0.09 & +16.7\% \\
\midrule
Qwen-2.5-72B & 44.4\% & 1.27 & 33.3\% & 70.5\% & 2.10 & 50.0\% & 52.5\% & 1.48 & 42.9\% & 21.0\% & 0.53 & 16.7\% \\
 \quad w/RAG & +3.2\% & +0.11 & +0.0\% & -27.9\% & -0.90 & +0.0\% & +6.2\% & +0.01 & +0.0\% & +9.6\% & +0.31 & +16.6\% \\
\midrule
MARG & 41.7\% & 1.33 & 50.0\% & 38.5\% & 1.38 & 33.3\% & 15.4\% & 0.38 & 28.6\% & 58.3\% & 1.67 & 50.0\% \\
 \quad w/ RAG & +16.6\% & +0.34 & +16.7\% & +30.7\% & +0.85 & +33.4\% & +30.8\% & +0.77 & +38.1\% & +16.7\% & +0.91 & +33.3\% \\
\bottomrule
\end{tabular}
}
\caption{
Human and automated evaluation results on Experimental Design in \oursone set.
}
\label{tab:sub_results_2}
\end{table*}

\begin{table*}[!t]
\centering
\resizebox{\linewidth}{!}{%
\small
\addtolength{\tabcolsep}{-0.1em}
\renewcommand{\arraystretch}{1.1}
\begin{tabular}{lcccccc}
\toprule
\multirow{3}{*}{\textbf{Systems}} & \multicolumn{3}{c}{\textbf{Limited Analysis}}  & \multicolumn{3}{c}{\textbf{Insufficient Metrics}}\\
\cmidrule(lr){2-4} \cmidrule(lr){5-7}
& \multicolumn{2}{c}{\textbf{Automated Eval.}}  & \multicolumn{1}{c}{\textbf{Human Eval.}} & \multicolumn{2}{c}{\textbf{Automated Eval.}}  & \multicolumn{1}{c}{\textbf{Human Eval.}}\\
\cmidrule(lr){2-3} \cmidrule(lr){4-4} \cmidrule(lr){5-6} \cmidrule(lr){7-7}
& Coarse & Fine (0-5) & Accuracy & Coarse & Fine (0-5) & Accuracy \\
\midrule
GPT-4o & 72.0\% & 1.88 & 84.6\% & 52.0\% & 1.36 & 50.0\% \\
GPT-4o w/ RAG & +28.0\% & +0.64 & +15.4\% & +8.0\% & +0.44 & +8.3\% \\
\midrule
GPT-4o-mini & 46.4\% & 1.10 & 46.2\% & 50.4\% & 1.43 & 41.7\% \\
GPT-4o-mini w/RAG & +15.2\% & +0.39 & +7.6\% & +0.0\% & +0.03 & +0.0\% \\
\midrule
Llama-3.3-70B & 59.0\% & 1.42 & 53.8\% & 47.9\% & 1.33 & 41.7\% \\
Llama-3.3-70B w/RAG & +20.2\% & +0.56 & +7.7\% & -6.0\% & -0.18 & -8.4\% \\
\midrule
Qwen-2.5-72B & 71.0\% & 1.72 & 61.5\% & 47.2\% & 1.26 & 41.7\% \\
Qwen-2.5-72B w/RAG & +21.0\% & +0.67 & +15.4\% & -4.0\% & -0.09 & +0.0\% \\
\midrule
MARG & 84.2\% & 2.58 & 84.6\% & 88.0\% & 2.32 & 75.0\% \\
MARG w/ RAG & +6.7\% & +0.10 & +15.4\% & +4.0\% & +0.12 & +8.3\% \\
\bottomrule
\end{tabular}
}
\caption{
Human and automated evaluation results on Result Analysis in \oursone set.
}
\label{tab:sub_results_3}
\end{table*}

\begin{table*}[!t]
\centering
\resizebox{\linewidth}{!}{%
\small
\addtolength{\tabcolsep}{-0.1em}
\renewcommand{\arraystretch}{1.1}
\begin{tabular}{lccccccccc}
\toprule
\multirow{3}{*}{\textbf{Systems}} & \multicolumn{3}{c}{\textbf{Limited Scope}}  & \multicolumn{3}{c}{\textbf{Irrelevant Citations}} & \multicolumn{3}{c}{\textbf{Inaccurate Description}}\\
\cmidrule(lr){2-4} \cmidrule(lr){5-7} \cmidrule(lr){8-10}
& \multicolumn{2}{c}{\textbf{Automated Eval.}}  & \multicolumn{1}{c}{\textbf{Human Eval.}} & \multicolumn{2}{c}{\textbf{Automated Eval.}}  & \multicolumn{1}{c}{\textbf{Human Eval.}} & \multicolumn{2}{c}{\textbf{Automated Eval.}}  & \multicolumn{1}{c}{\textbf{Human Eval.}}\\
\cmidrule(lr){2-3} \cmidrule(lr){4-4} \cmidrule(lr){5-6} \cmidrule(lr){7-7} \cmidrule(lr){8-9} \cmidrule(lr){10-10}
& Coarse & Fine (0-5) & Accuracy & Coarse & Fine (0-5) & Accuracy &  Coarse & Fine (0-5) & Accuracy \\
\midrule
GPT-4o & 100.0\% & 2.69 & 87.5\%  & 0.0\% & 0.00 & 0.0\% & 56.2\% & 1.25 & 50.0\% \\
GPT-4o w/ RAG & +0.0\% & +0.06 & +12.5\% & +0.0\% & +0.00 & +11.1\% & +12.6\% & +0.31 & +25.0\% \\
\midrule
GPT-4o-mini & 100.0\% & 2.72 & 75.0\% & 0.0\% & 0.00 & 0.0\% & 41.0\% & 1.00 & 37.5\% \\
GPT-4o-mini w/RAG & +0.0\% & +0.03 & +12.5\% & +1.2\% & +0.04 & +0.0\% & +6.0\% & +0.12 & +12.5\% \\
\midrule
Llama-3.3-70B & 97.4\% & 2.73 & 75.0\% & 2.5\% & 0.02 & 7.7\% & 15.4\% & 0.32 & 12.5\% \\
Llama-3.3-70B w/RAG & -15.9\% & -0.69 & +0.0\% & -2.5\% & -0.02 & -7.7\% & +0.6\% & +0.08 & +12.5\% \\
\midrule
Qwen-2.5-72B & 90.0\% & 2.35 & 87.5\% & 0.0\% & 0.00 & 0.0\% & 23.5\% & 0.48 & 25.0\% \\
Qwen-2.5-72B w/RAG & +6.3\% & +0.16 & +12.5\% & +0.0\% & +0.00 & +0.0\% & -4.5\% & -0.04 & +0.0\% \\
\midrule
MARG & 100.0\% & 3.00 & 100.0\% & 37.5\% & 0.94 & 22.2\% & 57.1\% & 1.29 & 37.5\% \\
MARG w/ RAG & +0.0\% & +0.06 & +0.0\% & +9.6\% & +0.24 & +11.1\% & +7.2\% & +0.28 & +25.0\% \\
\bottomrule
\end{tabular}
}
\caption{
Human and automated evaluation results on Literature Review in \oursone set.
}
\label{tab:sub_results_4}
\end{table*}

\begin{table}[!t]
\centering
\resizebox{\linewidth}{!}{%
\small
\addtolength{\tabcolsep}{-0.1em}
\renewcommand{\arraystretch}{1.1}
\begin{tabular}{lccc}
\toprule
\multirow{2}{*}{\textbf{Systems}} & \multicolumn{2}{c}{\textbf{Automated Eval.}}  & \multicolumn{1}{c}{\textbf{Human Eval.}}\\
\cmidrule(lr){2-3} \cmidrule(lr){4-4}
& Coarse & Fine (0-5) & Accuracy \\
\midrule
GPT-4o & 46.0\% & 1.12 & 35.9\% \\
GPT-4o w/ RAG & +10.0\% & +0.28 & +11.9\% \\
\midrule
GPT-4o-mini & 58.8\% & 1.39 & 43.9\% \\
GPT-4o-mini w/RAG & +2.8\% & +0.07 & +8.0\% \\
\midrule
Llama-3.3-70B & 55.8\% & 1.20 & 35.6\% \\
Llama-3.3-70B w/RAG & +3.1\% & +0.08 & +0.0\% \\
\midrule
Qwen-2.5-72B & 44.3\% & 1.01 & 19.9\% \\
Qwen-2.5-72B w/RAG & -2.1\% & -0.08 & +8.0\% \\
\midrule
MARG & 82.8\% & 1.94 & 72.1\% \\
MARG w/ RAG & +4.4\% & +0.05 & +11.5\% \\
\bottomrule
\end{tabular}
}
\caption{
Human and automated evaluation results on Methodology in \oursone set.
}
\label{tab:aspect_results_1}
\end{table}

\begin{table}[!t]
\centering
\resizebox{\linewidth}{!}{%
\small
\addtolength{\tabcolsep}{-0.1em}
\renewcommand{\arraystretch}{1.1}
\begin{tabular}{lccc}
\toprule
\multirow{2}{*}{\textbf{Systems}} & \multicolumn{2}{c}{\textbf{Automated Eval.}}  & \multicolumn{1}{c}{\textbf{Human Eval.}}\\
\cmidrule(lr){2-3} \cmidrule(lr){4-4}
& Coarse & Fine (0-5) & Accuracy \\
\midrule
GPT-4o & 47.9\% & 1.31 & 55.6\% \\
GPT-4o w/ RAG & +16.7\% & +0.54 & +22.2\% \\
\midrule
GPT-4o-mini & 42.3\% & 1.09 & 44.4\% \\
GPT-4o-mini w/RAG & +4.0\% & +0.17 & +5.6\% \\
\midrule
Llama-3.3-70B & 35.3\% & 1.01 & 37.3\% \\
Llama-3.3-70B w/RAG & +5.2\% & +0.12 & +15.9\% \\
\midrule
Qwen-2.5-72B & 47.1\% & 1.34 & 42.1\% \\
Qwen-2.5-72B w/RAG & -2.2\% & -0.12 & +0.0\% \\
\midrule
MARG & 38.5\% & 1.19 & 37.3\% \\
MARG w/ RAG & +23.7\% & +0.72 & +29.4\% \\
\bottomrule
\end{tabular}
}
\caption{
Human and automated evaluation results on Experimental Design in \oursone set.
}
\label{tab:aspect_results_2}
\end{table}

\begin{table}[!t]
\centering
\resizebox{\linewidth}{!}{%
\small
\addtolength{\tabcolsep}{-0.1em}
\renewcommand{\arraystretch}{1.1}
\begin{tabular}{lccc}
\toprule
\multirow{2}{*}{\textbf{Systems}} & \multicolumn{2}{c}{\textbf{Automated Eval.}}  & \multicolumn{1}{c}{\textbf{Human Eval.}}\\
\cmidrule(lr){2-3} \cmidrule(lr){4-4}
& Coarse & Fine (0-5) & Accuracy \\
\midrule
GPT-4o & 62.0\% & 1.62 & 67.3\% \\
GPT-4o w/ RAG & +18.0\% & +0.54 & +11.9\% \\
\midrule
GPT-4o-mini & 48.4\% & 1.27 & 43.9\% \\
GPT-4o-mini w/RAG & +7.6\% & +0.21 & +3.9\% \\
\midrule
Llama-3.3-70B & 53.5\% & 1.38 & 47.8\% \\
Llama-3.3-70B w/RAG & +7.1\% & +0.19 & -0.3\% \\
\midrule
Qwen-2.5-72B & 59.1\% & 1.49 & 51.6\% \\
Qwen-2.5-72B w/RAG & +8.5\% & +0.29 & +7.7\% \\
\midrule
MARG & 86.1\% & 2.45 & 79.8\% \\
MARG w/ RAG & +5.4\% & +0.11 & +11.9\% \\
\bottomrule
\end{tabular}
}
\caption{
Human and automated evaluation results on Result Analysis in \oursone set.
}
\label{tab:aspect_results_3}
\end{table}

\begin{table}[!t]
\centering
\resizebox{\linewidth}{!}{%
\small
\addtolength{\tabcolsep}{-0.1em}
\renewcommand{\arraystretch}{1.1}
\begin{tabular}{lccc}
\toprule
\multirow{2}{*}{\textbf{Systems}} & \multicolumn{2}{c}{\textbf{Automated Eval.}}  & \multicolumn{1}{c}{\textbf{Human Eval.}}\\
\cmidrule(lr){2-3} \cmidrule(lr){4-4}
& Coarse & Fine (0-5) & Accuracy \\
\midrule
GPT-4o & 52.1\% & 1.31 & 25.0\% \\
GPT-4o w/ RAG & +4.2\% & +0.13 & +18.1\% \\
\midrule
GPT-4o-mini & 47.0\% & 1.24 & 18.8\% \\
GPT-4o-mini w/RAG & +2.4\% & +0.06 & +6.2\% \\
\midrule
Llama-3.3-70B & 38.4\% & 1.02 & 10.1\% \\
Llama-3.3-70B w/RAG & -5.9\% & -0.21 & +2.4\% \\
\midrule
Qwen-2.5-72B & 37.8\% & 0.94 & 12.5\% \\
Qwen-2.5-72B w/RAG & +0.6\% & +0.04 & +0.0\% \\
\midrule
MARG & 64.9\% & 1.74 & 29.9\% \\
MARG w/ RAG & +5.6\% & +0.20 & +18.0\% \\
\bottomrule
\end{tabular}
}
\caption{
Human and automated evaluation results on Literature Review in \oursone set.
}
\label{tab:aspect_results_4}
\end{table}

\begin{table*}[!t]
\centering
\small
\addtolength{\tabcolsep}{0.1em}
\renewcommand{\arraystretch}{1.15}
\begin{tabular}{lp{1.3cm}p{1.3cm}p{1.3cm}p{1.3cm}p{1.3cm}p{1.3cm}}
\toprule
\multirow{2}{*}{\textbf{Systems}}  & \multicolumn{2}{c}{\textbf{Coarse.}} & \multicolumn{2}{c}{\textbf{Fine.}} & \multicolumn{2}{c}{\textbf{Accuracy}}\\
\cmidrule(lr){2-3} \cmidrule(lr){4-5} \cmidrule(lr){6-7}
 & Median & Variance   & Median & Variance  & Median & Variance\\
\midrule
GPT-4o & 52.0\% & 7.4\% & 1.25 & 0.60 & 50.0\% & 6.9\% \\
GPT-4o w/ RAG & 66.7\% & 9.3\% & 1.56 & 0.78 & 66.7\% & 8.4\% \\
\midrule
GPT-4o-mini & 50.4\% & 8.5\% & 1.10 & 0.64 & 41.7\% & 5.3\% \\
GPT-4o-mini w/ RAG & 60.0\% & 9.0\% & 1.46 & 0.69 & 50.0\% & 6.7\% \\
\midrule
Llama-3.3-70B & 47.9\% & 7.8\% & 1.01 & 0.64 & 28.6\% & 6.1\% \\
Llama-3.3-70B w/ RAG & 45.8\% & 7.1\% & 1.16 & 0.47 & 33.3\% & 6.3\% \\
\midrule
Qwen-2.5-72B & 45.5\% & 6.5\% & 1.26 & 0.50 & 33.3\% & 5.9\% \\
Qwen-2.5-72B w/ RAG & 43.2\% & 7.9\% & 1.17 & 0.56 & 33.3\% & 7.3\% \\
\midrule
MARG & 58.3\% & 7.2\% & 1.67 & 0.58 & 50.0\% & 6.5\% \\
MARG w/ RAG & 75.0\% & 3.6\% & 2.23 & 0.43 & 75.0\% & 3.8\% \\
\bottomrule
\end{tabular}
\caption{
The median and variance of the results across subtypes in \oursone set.
}
\label{tab:med_var_syn}
\end{table*}

\subsection{\ourstwo Experiments}

\begin{figure}[h]
\begin{tcolorbox}[colback=white, colframe=black!80!white, title=Overlap Evaluation, fontupper=\footnotesize, fonttitle=\footnotesize]
\texttt{[System Input]}: \vspace{2pt}\\
Compare the following pair of limitations: one generated and one from the ground truth.\\

Assess the degree of relatedness and the level of specificity of the generated limitation compared to the ground truth limitation. Start by providing a brief explanation for each category, and then assign a rating. Present your assessment in JSON format as follows:\\

\{\\
"relatedness\_reason": "< Provide a brief explanation of why you assessed the relatedness as you did>",\\
"relatedness": "<Choose one of the following options: 'none', 'weak', 'medium', 'high'>",\\
"specificity\_reason": "<Provide a brief explanation of why you assessed the specificity as you did>",\\
"specificity": "<Choose one of the following options: 'less', 'same', 'more'>"\\
\}\\\\

\texttt{[User Input]}: \vspace{2pt}\\
Ground truth limitation: \\
\{ground truth\}\\
Generated limitation: \\
\{generated limitation\} \\
\end{tcolorbox}
\caption{Prompt for measuring overlap in \ourstwo.}
\label{fig:prompt_eval3}
\end{figure}

\paragraph{Automated Evaluation Metrics} Given a set of generated limitations $C_{\text{gen}}$ and a set of ground truth limitations $C_{\text{gt}}$ for a paper, each generated limitation is paired with every ground truth limitation of the same aspect. GPT-4o assesses the degree of relatedness for each pair, categorizing them as "none," "weak," "medium," or "high." Pairs rated "medium" or "high" are counted as successful matches. Using the alignments between $C_{\text{gen}}$ and $C_{\text{gt}}$, we evaluate several metrics, as described below. We refer to MARG~\cite{d2024marg} and define directional intersection operators $\overset{\leftarrow{}}{\cap}$ and $\overset{\rightarrow{}}{\cap}$ to represent the set of aligned elements in the left or right operand, respectively. For example, $C_{\text{gen}}\overset{\leftarrow{}}{\cap}C_{\text{gt}}$ is the set of elements of $C_{\text{gen}}$ that align to any element in $C_{\text{gt}}$.

\begin{itemize}[leftmargin=*]
    \itemsep0em
    \item \textbf{Recall:} $\frac{|C_{\text{gen}}\overset{\rightarrow{}}{\cap}C_{\text{gt}}|}{C_{\text{gt}}}$, the fraction of real reviewer comments that are aligned to any generated limitations.
    \item \textbf{Precision:} $\frac{|C_{\text{gen}}\overset{\leftarrow{}}{\cap}C_{\text{gt}}|}{C_{\text{gen}}}$, the fraction of generated limitations that are aligned to any ground truth limitation.
    \item \textbf{(Pseudo-)Jaccard:} The Jaccard index is a commonly used measure of set overlap. Let $Intersection = \frac{|C_{\text{gen}}\overset{\leftarrow{}}{\cap}C_{\text{gt}}| + |C_{\text{gen}}\overset{\rightarrow{}}{\cap}C_{\text{gt}}|}{2}$; then the Jaccard Index is $\frac{intersection}{|C_{\text{gen}}| + |C_{\text{gt}}| - intersection}$.
\end{itemize}

We adopt a macro-averaging approach at the individual paper level. We generate several limitations for each paper in the test set and compare them with the corresponding human-written limitations, calculating the relevant metrics for each comparison. These metrics are then averaged across all the papers to produce a single aggregated value for each metric.

\paragraph{Result Analysis} 
\autoref{tab:aspect_results_5} to \autoref{tab:aspect_results_8} show the detailed result for all the aspects in \ourstwo.

Overall, LLMs exhibit higher overlap and better quality in generating limitations related to experimental design compared to human reviewers. This may be because experimental design often receives the most feedback from human reviewers, providing a clearer reference in the automated evaluation. Also, limitations in experimental design tend to be more structured and objective, which makes it easier for LLMs to identify and refine issues. 

In contrast, LLMs perform the weakest in identifying limitations within the Literature Review aspect, which is consistent with the results observed in our \oursone. And RAG proves to be most helpful in this aspect. By retrieving and incorporating relevant papers, RAG helps the model identify missing references, overlooked methodologies, or underexplored areas, leading to more comprehensive and informed limitations.

\begin{table*}[!t]
\centering
\small
\addtolength{\tabcolsep}{0.1em}
\renewcommand{\arraystretch}{1.15}
\begin{tabular}{lp{1.3cm}p{1.3cm}p{1.3cm}p{1.3cm}p{1.3cm}p{1.3cm}p{1.3cm}}
\toprule
\multirow{2}{*}{\textbf{Systems}}  & \multicolumn{4}{c}{\textbf{Automated Evaluation}} & \multicolumn{3}{c}{\textbf{Human Evaluation (1-5)}}\\
\cmidrule(lr){2-5} \cmidrule(lr){6-8}
 & Recall & Precision & Jaccard & Fine (0-5) & Faith. & Sound. & Import.\\
\midrule
GPT-4o & 46.7\% & 20.9\% & 17.1\% & 0.43 & 3.14 & 2.82 & 3.62 \\
GPT-4o w/ RAG & +2.1\% & +2.8\% & +2.6\% & +0.12 & +0.66 & +1.17 & +0.40 \\
\midrule
GPT-4o-mini & 42.2\% & 19.4\% & 16.5\% & 0.38 & 2.92 & 2.66 & 3.10 \\
GPT-4o-mini w/ RAG & +5.0\% & +1.3\% & +1.3\% & +0.01 & +0.30 & +1.20 & +0.56 \\
\midrule
Llama-3.3-70B & 56.7\% & 19.9\% & 17.3\% & 0.42 & 3.04 & 2.78 & 3.24 \\
Llama-3.3-70B w/ RAG & +2.8\% & -0.1\% & -0.4\% & +0.03 & +0.28 & +1.09 & +0.20 \\
\midrule
Qwen-2.5-72B & 23.9\% & 21.6\% & 14.2\% & 0.49 & 2.52 & 2.78 & 2.94 \\
Qwen-2.5-72B w/ RAG & +5.2\% & +0.9\% & +1.0\% & +0.11 & +0.30 & +0.95 & +0.16 \\
\midrule
MARG & 51.9\% & 14.9\% & 12.4\% & 0.64 & 3.21 & 2.67 & 3.56 \\
MARG w/ RAG & +1.9\% & -1.1\% & -0.7\% & +0.24 & +0.97 & +1.41 & +0.33 \\
\bottomrule
\end{tabular}
\caption{
Human and automated evaluation results on Methodology in \ourstwo set.
}
\label{tab:aspect_results_5}
\end{table*}

\begin{table*}[!t]
\centering
\small
\addtolength{\tabcolsep}{0.1em}
\renewcommand{\arraystretch}{1.15}
\begin{tabular}{lp{1.3cm}p{1.3cm}p{1.3cm}p{1.3cm}p{1.3cm}p{1.3cm}p{1.3cm}}
\toprule
\multirow{2}{*}{\textbf{Systems}}  & \multicolumn{4}{c}{\textbf{Automated Evaluation}} & \multicolumn{3}{c}{\textbf{Human Evaluation (1-5)}}\\
\cmidrule(lr){2-5} \cmidrule(lr){6-8}
 & Recall & Precision & Jaccard & Fine (0-5) & Faith. & Sound. & Import.\\
\midrule
GPT-4o & 61.4\% & 34.1\% & 28.0\% & 0.70 & 3.70 & 3.41 & 4.12 \\
GPT-4o w/ RAG & +4.5\% & +2.6\% & +3.2\% & +0.13 & +0.68 & +1.14 & +0.47 \\
\midrule
GPT-4o-mini & 56.3\% & 33.3\% & 27.6\% & 0.70 & 3.76 & 3.40 & 3.66 \\
GPT-4o-mini w/ RAG & -1.4\% & -1.5\% & -1.6\% & -0.03 & +0.28 & +0.77 & -0.17 \\
\midrule
Llama-3.3-70B & 66.7\% & 31.9\% & 27.6\% & 0.63 & 3.28 & 3.63 & 3.56 \\
Llama-3.3-70B w/ RAG & +2.6\% & +0.0\% & +0.1\% & +0.06 & +0.19 & +0.57 & -0.21 \\
\midrule
Qwen-2.5-72B & 36.1\% & 42.0\% & 24.5\% & 0.94 & 3.79 & 3.63 & 3.67 \\
Qwen-2.5-72B w/ RAG & +0.4\% & +1.0\% & +1.5\% & +0.12 & +0.10 & +0.19 & +0.25 \\
\midrule
MARG & 54.9\% & 22.3\% & 18.8\% & 0.92 & 3.56 & 3.54 & 3.89 \\
MARG w/ RAG & +1.5\% & +3.8\% & +2.6\% & +0.21 & +0.49 & +0.70 & +0.34 \\
\bottomrule
\end{tabular}
\caption{
Human and automated evaluation results on Experiment Design in \ourstwo set.
}
\label{tab:aspect_results_6}
\end{table*}

\begin{table*}[!t]
\centering
\small
\addtolength{\tabcolsep}{0.1em}
\renewcommand{\arraystretch}{1.15}
\begin{tabular}{lp{1.3cm}p{1.3cm}p{1.3cm}p{1.3cm}p{1.3cm}p{1.3cm}p{1.3cm}}
\toprule
\multirow{2}{*}{\textbf{Systems}}  & \multicolumn{4}{c}{\textbf{Automated Evaluation}} & \multicolumn{3}{c}{\textbf{Human Evaluation (1-5)}}\\
\cmidrule(lr){2-5} \cmidrule(lr){6-8}
 & Recall & Precision & Jaccard & Fine (0-5) & Faith. & Sound. & Import.\\
\midrule
GPT-4o & 45.1\% & 14.8\% & 12.6\% & 0.36 & 3.21 & 2.73 & 3.41 \\
GPT-4o w/ RAG & -1.5\% & +2.8\% & +1.8\% & +0.09 & +0.05 & +1.15 & +0.40 \\
\midrule
GPT-4o-mini & 41.1\% & 15.5\% & 12.7\% & 0.32 & 3.07 & 2.93 & 2.87 \\
GPT-4o-mini w/ RAG & +3.3\% & +2.0\% & +1.5\% & +0.05 & +0.46 & +0.00 & +1.14 \\
\midrule
Llama-3.3-70B & 49.1\% & 15.5\% & 12.9\% & 0.32 & 3.00 & 2.73 & 2.81 \\
Llama-3.3-70B w/ RAG & +4.5\% & +0.6\% & +0.6\% & +0.03 & +0.38 & +0.12 & +0.51 \\
\midrule
Qwen-2.5-72B & 27.7\% & 18.9\% & 12.9\% & 0.48 & 2.86 & 2.81 & 2.91 \\
Qwen-2.5-72B w/ RAG & +1.6\% & +2.2\% & +1.2\% & +0.13 & +0.32 & +0.14 & +0.48 \\
\midrule
MARG & 59.4\% & 12.5\% & 11.1\% & 0.45 & 3.80 & 2.84 & 3.95 \\
MARG w/ RAG & +4.2\% & +3.3\% & +2.2\% & +0.16 & +0.42 & +1.08 & +0.19 \\

\bottomrule
\end{tabular}
\caption{
Human and automated evaluation results on Result Analysis in \ourstwo set.
}
\label{tab:aspect_results_7}
\end{table*}

\begin{table*}[!t]
\centering
\small
\addtolength{\tabcolsep}{0.1em}
\renewcommand{\arraystretch}{1.15}
\begin{tabular}{lp{1.3cm}p{1.3cm}p{1.3cm}p{1.3cm}p{1.3cm}p{1.3cm}p{1.3cm}}
\toprule
\multirow{2}{*}{\textbf{Systems}}  & \multicolumn{4}{c}{\textbf{Automated Evaluation}} & \multicolumn{3}{c}{\textbf{Human Evaluation (1-5)}}\\
\cmidrule(lr){2-5} \cmidrule(lr){6-8}
 & Recall & Precision & Jaccard & Fine (0-5) & Faith. & Sound. & Import.\\
\midrule
GPT-4o & 31.8\% & 6.6\% & 5.8\% & 0.18 & 2.71 & 2.38 & 2.81 \\
GPT-4o w/ RAG & +10.9\% & +5.2\% & +4.1\% & +0.17 & +0.56 & +1.10 & +1.15 \\
\midrule
GPT-4o-mini & 21.1\% & 6.3\% & 5.2\% & 0.14 & 2.37 & 2.14 & 2.24 \\
GPT-4o-mini w/ RAG & +7.5\% & +1.0\% & +1.1\% & +0.05 & +0.09 & +1.11 & +0.61 \\
\midrule
Llama-3.3-70B & 39.5\% & 8.6\% & 7.5\% & 0.20 & 2.60 & 2.25 & 2.59 \\
Llama-3.3-70B w/ RAG & +2.8\% & +0.1\% & +0.1\% & +0.01 & +0.07 & +1.04 & +0.34 \\
\midrule
Qwen-2.5-72B & 16.9\% & 7.8\% & 6.0\% & 0.22 & 2.46 & 2.24 & 2.22 \\
Qwen-2.5-72B w/ RAG & -0.6\% & +1.3\% & +0.2\% & +0.06 & +0.18 & +0.11 & +0.50 \\
\midrule
MARG & 65.9\% & 22.1\% & 18.4\% & 0.62 & 3.84 & 3.71 & 3.72 \\
MARG w/ RAG & +4.3\% & +5.3\% & +5.9\% & +0.36 & +0.19 & +0.71 & +0.84 \\
\bottomrule
\end{tabular}

\caption{
Human and automated evaluation results on Literature Review in \ourstwo set.
}
\label{tab:aspect_results_8}
\end{table*}

\begin{table}[!t]
\centering
\resizebox{\linewidth}{!}{%
\small
\addtolength{\tabcolsep}{-0.1em}
\renewcommand{\arraystretch}{1.1}
\begin{tabular}{lccccc}
\toprule
\multirow{2}{*}{\textbf{Systems}} & \multicolumn{2}{c}{\textbf{Automated Eval.}}  & \multicolumn{3}{c}{\textbf{Human Eval. (1-5)}}\\
\cmidrule(lr){2-3} \cmidrule(lr){4-6}
& Jaccard & Fine.(0-5) & Faith. & Sound. & Import. \\
\midrule
GPT-4o & 14.9\% & 0.39 & 3.17 & 2.78 & 3.52 \\
\quad w/ RAG & +2.2\% & +0.11 & +0.37 & +1.15 & +0.47 \\
\midrule
GPT-4o-mini & 14.6\% & 0.35 & 3.00 & 2.80 & 2.99 \\
\quad w/ RAG & +1.4\% & +0.03 & +0.37 & +0.76 & +0.58 \\
\midrule
Llama-3.3-70B & 15.1\% & 0.37 & 3.02 & 2.75 & 3.02 \\
\quad w/ RAG & +0.1\% & +0.04 & +0.33 & +0.83 & +0.32 \\
\midrule
Qwen-2.5-72B & 13.6\% & 0.48 & 2.69 & 2.79 & 2.93 \\
\quad w/ RAG & +1.1\% & +0.12 & +0.31 & +0.55 & +0.32 \\
\midrule
MARG & 15.4\% & 0.63 & 3.68 & 3.19 & 3.81 \\
\quad w/ RAG & +1.9\% & +0.30 & +0.44 & +0.97 & +0.38 \\
\bottomrule
\end{tabular}
}

\caption{
The median of the results across aspects in \ourstwo set.
}
\label{tab:med_human}
\end{table}

\begin{table}[!t]
\centering
\resizebox{\linewidth}{!}{%
\small
\addtolength{\tabcolsep}{-0.1em}
\renewcommand{\arraystretch}{1.1}
\begin{tabular}{lccccc}
\toprule
\multirow{2}{*}{\textbf{Systems}} & \multicolumn{2}{c}{\textbf{Automated Eval.}}  & \multicolumn{3}{c}{\textbf{Human Eval. (1-5)}}\\
\cmidrule(lr){2-3} \cmidrule(lr){4-6}
& Jaccard & Fine.(0-5) & Faith. & Sound. & Import. \\
\midrule
GPT-4o & 0.9\% & 0.05 & 0.16 & 0.18 & 0.29 \\
\quad w/ RAG & 0.8\% & 0.04 & 0.28 & 0.20 & 0.12 \\
\midrule
GPT-4o-mini & 0.9\% & 0.06 & 0.33 & 0.28 & 0.34 \\
\quad w/ RAG & 0.7\% & 0.04 & 0.44 & 0.32 & 0.24 \\
\midrule
Llama-3.3-70B & 0.7\% & 0.03 & 0.08 & 0.33 & 0.19 \\
\quad w/ RAG & 0.7\% & 0.04 & 0.13 & 0.36 & 0.05 \\
\midrule
Qwen-2.5-72B & 0.6\% & 0.09 & 0.38 & 0.33 & 0.35 \\
\quad w/ RAG & 0.7\% & 0.10 & 0.31 & 0.48 & 0.25 \\
\midrule
MARG & 0.2\% & 0.04 & 0.08 & 0.26 & 0.03 \\
\quad w/ RAG & 0.4\% & 0.05 & 0.01 & 0.05 & 0.08 \\
\bottomrule
\end{tabular}
}

\caption{
The variance of the results across aspects in \ourstwo set.
}
\label{tab:var_human}
\end{table}
\clearpage
\subsection{Case study}\label{app:case}
\begin{figure*}[!t]
\centering
\includegraphics[width=1\textwidth]{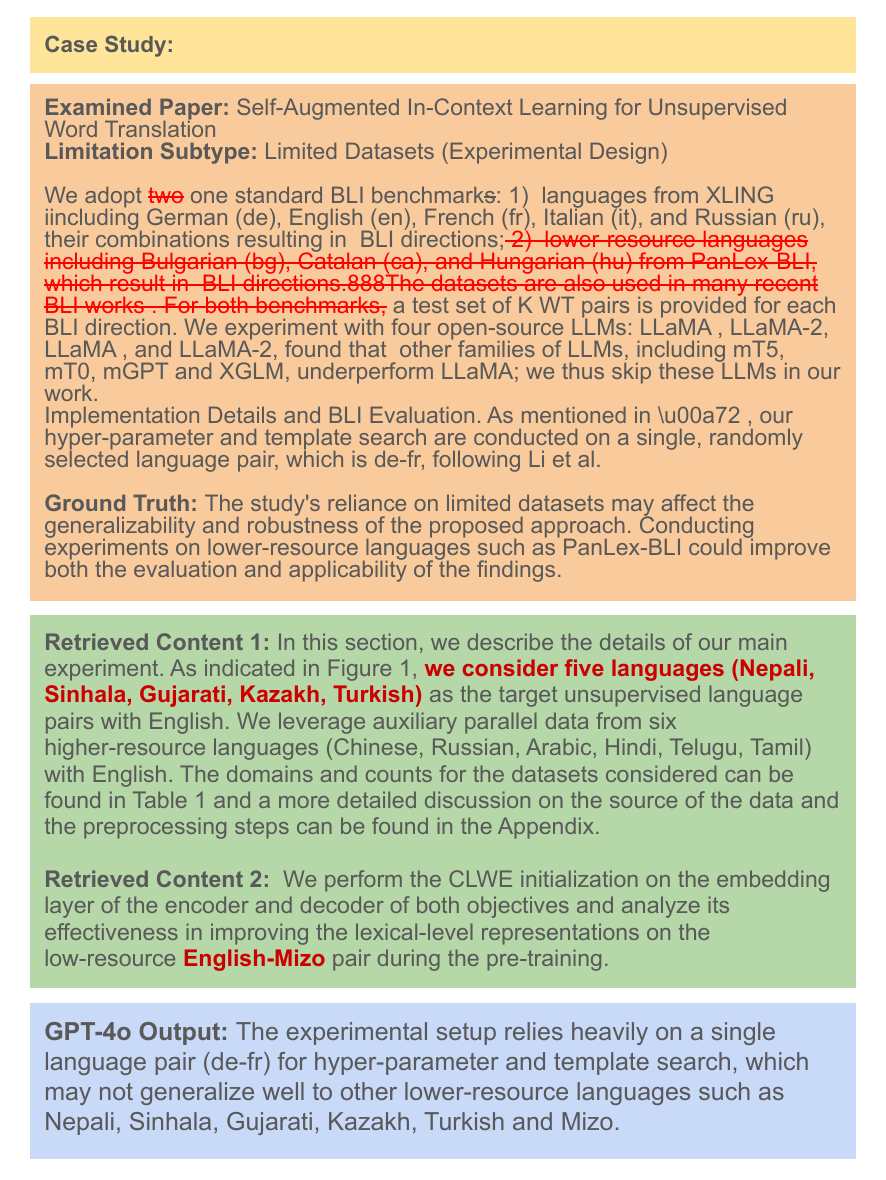}
\caption{
An example of GPT-4o w/ RAG generated limitation in \oursone.
}
\label{fig:case_1}
\end{figure*}

\begin{figure*}[!t]
\centering
\includegraphics[width=1\textwidth]{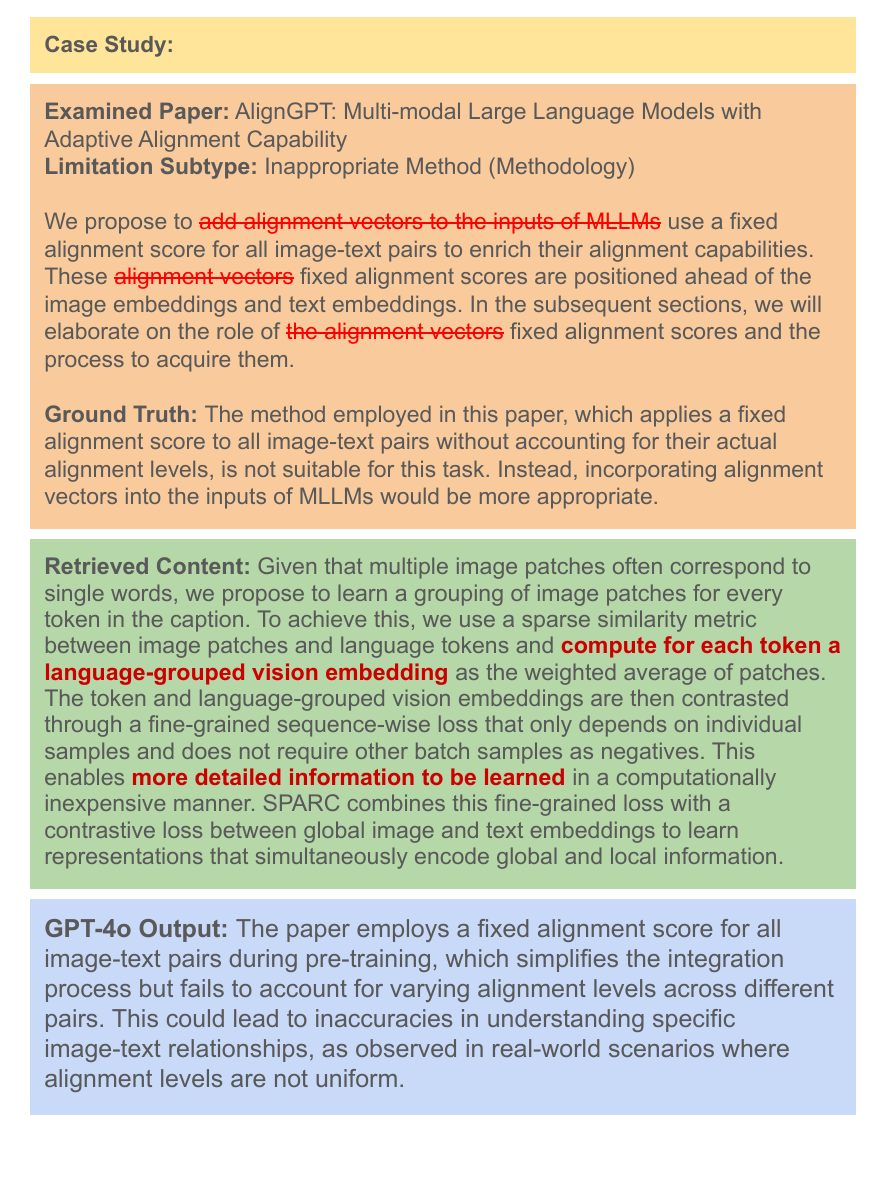}
\caption{
An example of GPT-4o w/ RAG generated limitation in \oursone.
}
\label{fig:case_2}
\end{figure*}

\begin{figure*}[!t]
\centering
\includegraphics[width=1\textwidth]{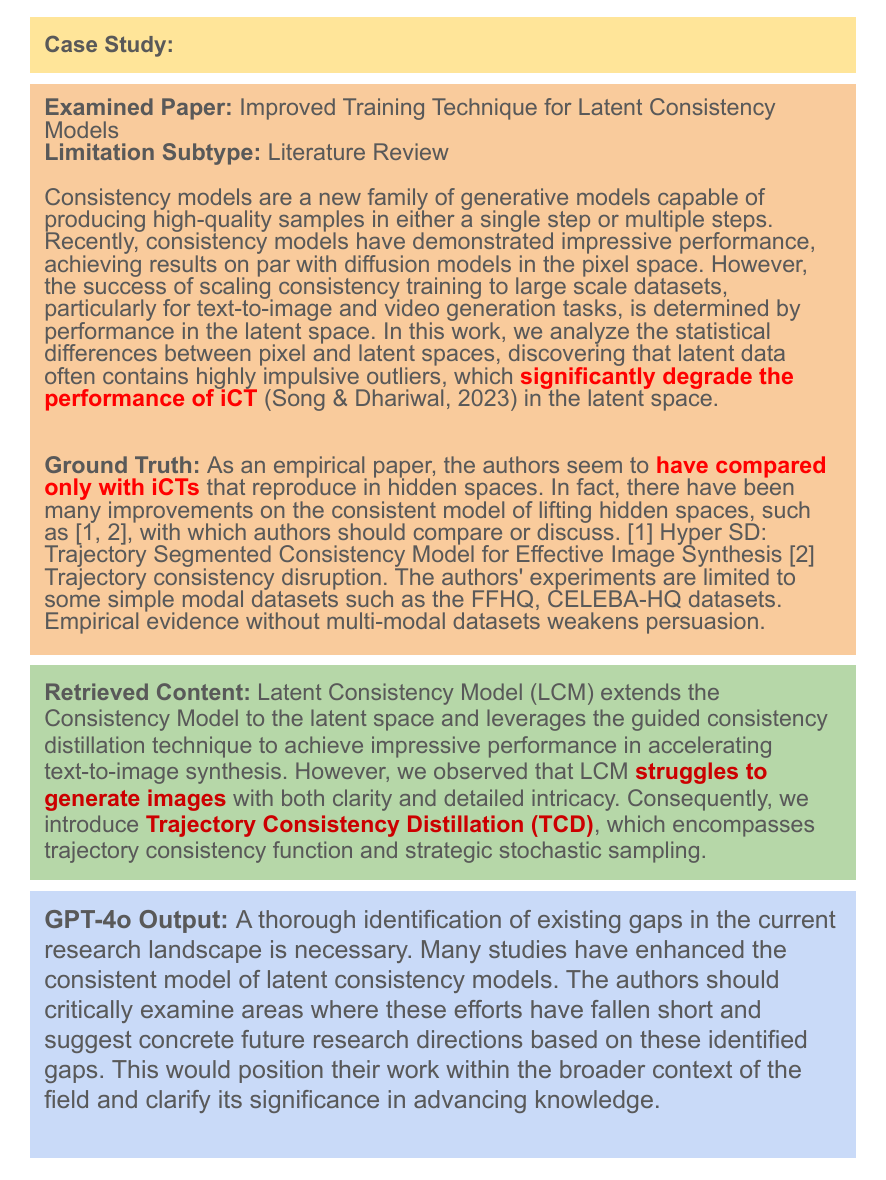}
\caption{
An example of GPT-4o w/ RAG generated limitation in \ourstwo.
}
\label{fig:case_3}
\end{figure*}

\begin{figure*}[!t]
\centering
\includegraphics[width=1\textwidth]{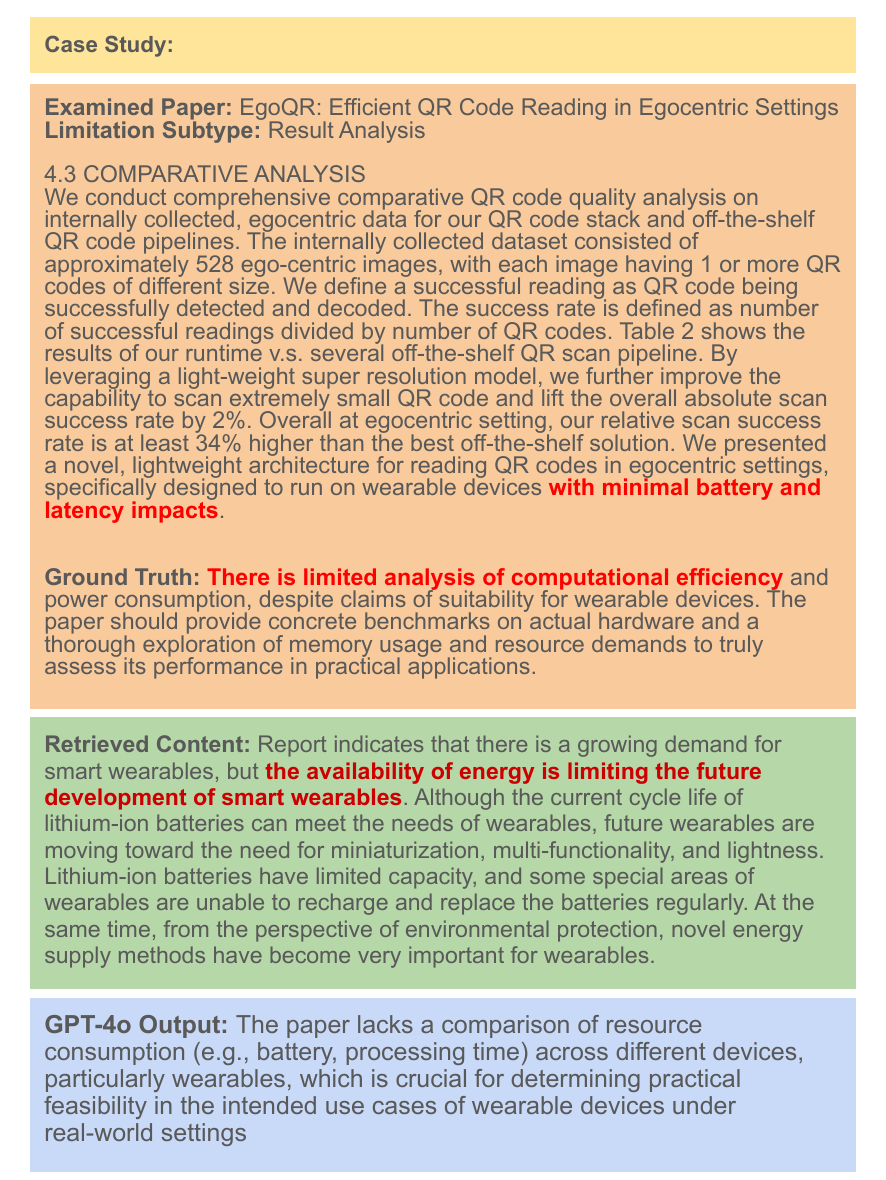}
\caption{
An example of GPT-4o w/ RAG generated limitation in \ourstwo.
}
\label{fig:case_4}
\end{figure*}

\end{document}